\documentclass[11pt]{article}

\usepackage{epsfig,amsmath,latexsym,amssymb}
\usepackage{graphicx}
\usepackage{lscape}
\usepackage{picture, eso-pic, tikz} 
\usepackage{dsfont}
\usepackage{listings}
\usepackage{changes}
\usepackage{hyperref}

\renewcommand{\baselinestretch}{1.1}
\def\R{{\mathbb R}}  
\def\N{{\mathbb N}}  

\newcommand{\Remm}[1]{}

\newtheorem{model ass}[theo]{Model Assumptions}

\numberwithin{equation}{section}

\definecolor{MyGray}{rgb}{0.92,0.92,0.92}


\newcommand{\bl}[1]{\textcolor{blue}{{#1}}}

\definecolor{British racing}{rgb}{0.0, 0.5, 0.0}
\def\bx{\boldsymbol{x}}
\def\be{\boldsymbol{e}}
\def\bz{\boldsymbol{z}}

\def\b0{\boldsymbol{0}}
\def\ba{\boldsymbol{a}}

\def\b0{\boldsymbol{0}}
\def\ba{\boldsymbol{a}}

\def\bc{\boldsymbol{c}}
\def\bk{\boldsymbol{k}}
\def\bq{\boldsymbol{q}}
\def\bv{\boldsymbol{v}}
\def\bbb{\boldsymbol{b}}

\begin{document}
\author{Ronald Richman\footnote{Old Mutual Insure and University of the Witwatersrand, ronald.richman@ominsure.co.za}
\and Salvatore Scognamiglio \footnote{Department of Management and Quantitative Sciences, University of Naples ``Parthenope",\newline salvatore.scognamiglio@uniparthenope.it}
\and Mario V.~W\"uthrich\footnote{RiskLab, Department of Mathematics, ETH Zurich,
mario.wuethrich@math.ethz.ch}}

\date{Version of \today}
\title{The Credibility Transformer}
\maketitle

\begin{abstract}
\noindent  
Inspired by the large success of Transformers in Large Language Models, these architectures are increasingly applied to tabular data. This is achieved by embedding tabular data into low-dimensional Euclidean spaces resulting in similar structures as time-series data. We introduce a novel credibility mechanism to this Transformer architecture. This credibility mechanism is based on a special token that
should be seen as an encoder that consists of a credibility weighted average of prior information and observation based information. We demonstrate that this novel credibility mechanism is very beneficial to stabilize training, and our
Credibility Transformer leads to predictive models that are superior to state-of-the-art deep learning models.

\bigskip


\end{abstract}

\section{Introduction}
Feed-forward neural networks (FNNs) provide state-of-the-art deep learning regression models for actuarial pricing. FNNs can be seen as extensions of generalized linear models (GLMs),
taking covariates as inputs to these FNNs, feature-engineering these covariates through several hidden FNN layers, and then using these feature-engineered covariates as inputs to a GLM. Advantages of FNNs over classical GLMs are that they are able to find functional forms and interactions in the covariates that cannot easily be captured by GLMs, and which typically require the modeler to have specific deeper insights into the data generation process. Since these specific deeper insights are not always readily available, FNNs may support the modeler in finding such structure and insight.

Taking inspiration from the recent huge success of large language models (LLMs), the natural
question arises whether there are network architectures other than FNNs that share more similarity with LLMs and which can further improve predictive performance of neural networks in actuarial pricing. LLMs are based on the Transformer architecture which has been invented by Vaswani et al.~\cite{Vaswani}. The Transformer architecture is based on attention layers which are special
network modules that allow covariate components to communicate with each other. Specifically, one may think of each covariate component receiving a so-called query and a key, and the attention mechanism tries to find queries and keys of different covariate components that match in order to send forward a positive or negative signal. For example, in the case of pricing motor insurance for car drivers, young drivers may have a key labeled `risky', and car brands may have queries, for instance, sports cars trying to find the age group of risky drivers. Having a match of a query and a key then leads to an increase of the expected claim frequency, which accounts for a corresponding
interaction of these two covariates in the regression function. This idea of keys, queries and values, which is central to Transformer models, originates in the information retrieval literature, where a search term (the query) is matched to relevant documents (the keys), the contents of which provide the information the user is looking for (the values).

Transformer architectures have been introduced for natural language data (i.e., time-series data); see Vaswani et al.~\cite{Vaswani}. A main question is whether there is a similar beneficial way to use
Transformers and attention layers on tabular input data, which lack the same time-series structure. The main tool for applying Transformers to tabular data is the so-called entity embedding mechanism. Embedding all covariates into a low-dimensional Euclidean space, one receives embedded tabular covariates that share similar features as time-series data. The first attempt at building this type of model is found in Huang et al.~\cite{Huang}. These authors proposed a model embedding only categorical components of tabular data, applying a Transformer to these, and then concatenating the Transformer outputs with the numerical input variables for further processing this input information by a FNN. This TabTransformer model of Huang et al.~\cite{Huang} has been applied by Kuo--Richman \cite{Kuo}, who found that this model provides a small advantage over FNN architectures for predicting severity of flood claims. A more comprehensive approach has recently been proposed by Gorshniy et al.~\cite{Gorishniy}, the feature tokenizer (FT) Transformer architecture, which embeds numerical components of tabular data as well. This proposal has been considered in the actuarial literature by Brauer \cite{Brauer}. Thus, technically there is no difficulty in using FT-Transformers on tabular data. However, in view of  the numerical results obtained in Brauer \cite{Brauer}, there does not seem to be any specific benefit in using a FT-Transformer architecture 
for tabular data in terms of predictive performance. This is precisely the starting point of this research. Namely, we are going to modify the FT-Transformer architecture by a novel credibility mechanism. Using this additional credibility mechanism we find Transformer-based network architectures that outperform the networks that have been exploited so far on a specific motor insurance dataset. This gives clear evidence that Transformers can also be very beneficial on tabular input data
if network architectures are designed in a sophisticated manner.

Our main idea is to add a special token to the classical Transformer architecture, inspired by the Bidirectional Encoder Representations
from Transformers (BERT) architecture of Devlin et al.~\cite{Devlin}. This special token (called the CLS token by Devlin et al.~\cite{Devlin}) is used to encode the covariate information gained from the attention mechanism in a lower dimensional representation. Through the training process described later, one is able think of this token as playing the role of prior information consisting of the portfolio mean in a Bayesian credibility sense. This prior information-based predictor is then combined in a linear credibility fashion (in the abstract embedding space of the model) with a second predictor that plays the role of observed information in Bayesian credibility theory. Specifically, we consider a credibility-based average of the two sets of information, and this provides us with a similar credibility structure as, e.g., the classical linear credibility formula of B\"uhlmann--Straub \cite{BS}. This motivates the use of the term {\it Credibility Transformer} for our proposal. In other words, the Credibility Transformer does not only try to find matches between keys and queries, but it also weighs this information by a credibility mechanism. As noted above, in this work, we make two main connections to the linear credibility formula: first, when training the model, we force the special CLS token to provide prior information to the model using a credibility factor, and second, we interpret the attention mechanism within the Transformer as a linear credibility formula between prior information contained in the CLS token and the rest of the covariate information.

Moreover, we exploit special fitting strategies of these Credibility Transformers since training Transformer architectures needs a sort of tempered learning to not get trapped in too early stopping decisions. For this we adapt the {\tt NormFormer} proposal of Shleifer et al.~\cite{NormFormer}, which applies a special Transformer pre-training that can cope with different gradient magnitudes in stochastic gradient descent training. We verify that this proposal of Shleifer et al.~\cite{NormFormer} is also beneficial in our Credibility Transformer architecture resulting in superior performance compared to plain-vanilla trained architectures. Building on this initial exploration, we then augment the Credibility Transformer with several advances from the LLM and deep learning literature, producing a deep and multi-head attention version of the initial Credibility Transformer architecture. 
Furthermore, we implement the concept of Gated Linear Units (GLU) of 
Dauphin et al.~~\cite{Dauphin17a} to select more important covariate
components, and we exploit the Piecewise Linear Encoding (PLE) of
Gorshniy et al.~\cite{Gorishniy} which should be thought of as
a more informative embedding, especially adapted for numerical covariates, than its one-hot encoded counterpart.
It is more informative because PLE preserves the ordinal relation
of continuous covariates, while enabling subsequent network layers to produce a multidimensional embedding for the numerical data. This {\it improved deep Credibility Transformer} equips us with regression models that have an excellent predictive performance. Remarkably, we show that the Credibility Transformer approach can improve a state-of-the-art deep Transformer model applied to a non-life pricing problem. Finally, we examine the explainability of the Credibility Transformer approach by exploring a fitted model.

\bigskip

{\bf Organization of this manuscript.} This paper is organized as follows. Section \ref{Model architecture section} describes the architecture of the Credibility Transformer in detail, including the input tokenization process, the Transformer layer with its attention mechanism, and our novel credibility mechanism. Section \ref{Real Data} presents a real data example using the French motor third party liability (MTPL) claims counts dataset, demonstrating the implementation and performance of the Credibility Transformer. Section \ref{Improving CT} explores improvements to the Credibility Transformer, incorporating insights from LLMs and recent advances in deep learning. Section \ref{Exploring CT} discusses the insights that can be gained from a fitted Credibility Transformer model. Finally, Section \ref{Conclusions} concludes by summarizing our contribution and findings and discusses potential future research directions.

\section{The Credibility Transformer}
\label{Model architecture section}
This section describes the architecture of the Credibility Transformer. The reader should have the classical Transformer architecture of Vaswani et al.~\cite{Vaswani} in mind, with one essential difference, namely, the entire set of relevant covariate information is going to be encoded into additional classify (CLS) tokens, and these CLS tokens are going to be combined with a credibility mechanism. For this we extend the classical Transformer architecture of Vaswani et al.~\cite{Vaswani} by CLS tokens. CLS tokens have been introduced by Devlin et al.~\cite{Devlin} in the language model Bidirectional Encoder Representations from Transformers (BERT).
In BERT, these tokens are used to summarize the probability of one sentence following an other one, therefore they have used the term `classify token'. In our architecture, this terminology may be a bit misleading because our CLS tokens will just be real numbers not relating to any classification 
probabilities, but they will rather present biases in regression models. Nevertheless, we keep the term CLS token to emphasize its structural analogy to BERT.

\subsection{Construction of the input tensor}
We first describe the input pre-processing. This input pre-processing tokenizes all
input variables (features, covariates) by embedding them into a low dimensional Euclidean space. These embeddings are complemented by positional embeddings and CLS tokens. We describe this in detail in the next three subsections. The main difference to the classical Transformer of Vaswani et al.~\cite{Vaswani} concerns the last step of adding the CLS tokens for encoding the information from
the attention mechanism, this feature is described in Section \ref{section Classify token}, below.

\subsubsection{Feature tokenizer}
We start by describing the tokenization of the covariates (input features). Assume we have $T_1$ categorical covariates $(x_t)_{t=1}^{T_1}$ and $T_2$ continuous covariates $(x_t)_{t=T_1+1}^{T} \in \R^{T_2}$; we set $T=T_1+T_2$ for the total number of covariate components in input $\bx=(x_t)_{t=1}^T$.

We bring these categorical and continuous input variables into the same structure by applying entity embedding $(\be^{\rm EE}_t)_{t=1}^{T_1}$  to the categorical variables and FNN embedding $(\be^{\rm FNN}_t)_{t=T_1+1}^T$ to the continuous variables; this pre-processing
step is called {\it feature tokenizer}. Entity embedding has been introduced in the context of natural language processing (NLP) by Br\'ebisson et al.~\cite{Brebisson} and Guo--Berkhahn \cite{Guo}, 
and it has been introduced to the actuarial community by Richman \cite{Richman2020a, Richman2020b} and Delong--Kozak \cite{Delong}. The main purpose of entity embedding is to represent (high-cardinality) nominal features by low dimensional Euclidean embeddings such that proximity in these low dimensional Euclidean space means similarity w.r.t.~the prediction task at hand. Additionally, we also embed the continuous input variables into the same low dimensional Euclidean space such that
all input variables have the same structure. This approach has been considered in Gorshniy et al.~\cite{Gorishniy}, and it is used by Brauer \cite{Brauer} in the actuarial literature.

The embeddings are defined as follows. For the entity embeddings of the categorical covariates we select mappings
\begin{equation*}
\be^{\rm EE}_t : \{a_1,\ldots, a_{n_t}\} \to \R^b,
\qquad 
x_t  \mapsto  \be^{\rm EE}_t(x_t),
\end{equation*}
where $a_1,\ldots, a_{n_t}$ are the different levels of the $t$-th
categorical covariate $x_t$, and $b \in \N$ is the fixed, pre-chosen embedding dimension; this embedding dimension $b$ is a hyperparameter that needs to be selected by the modeler. Importantly, for our Credibility Transformer all embeddings must have the same embedding dimension $b$ (possibly enlarged, as we do later, by concatenating other vectors to these embeddings); this embedding dimension $b$ is called the \textit{model dimension} of the Transformer model.

Each of these categorical entity embeddings involves $b n_t$ embedding weights (parameters), thus, we have $\sum_{t=1}^{T_1} b n_t$ parameters in total from the embedding of the categorical covariates $(x_t)_{t=1}^{T_1}$.

For the continuous input variables we select fully-connected FNN architectures $\bz_t^{(2:1)}=\bz_t^{(2)}\circ \bz_t^{(1)}$ of depth 2 being composed of two FNN layers
\begin{equation}\label{numerical embeddings layers}
\bz_t^{(1)}: \R \to \R^b \qquad \text{ and } \qquad 
\bz_t^{(2)}: \R^b \to \R^b.
\end{equation}
We select such a FNN architecture 
$\bz_t^{(2:1)}$ for each continuous covariate component
$x_t$,  $T_1+1\le t \le T$.
Each of these continuous FNN embeddings 
$\bz_t^{(2:1)}$ involves $2b + b(b+1)$ network weights (including the bias parameters). This gives us $T_2 (2b + b(b+1))$ 
parameters for the embeddings of the continuous covariates
$(x_t)_{t=T_1+1}^{T}$; for more technical details and the
notation used for FNNs we refer to W\"uthrich--Merz \cite[Chapter 7]{WM2023}.

Concatenating and reshaping these categorical and continuous covariate embeddings
equips us with the {\it raw input tensor} 
\begin{equation*}
\bx=(x_t)_{t=1}^T ~\mapsto ~ \bx^\circ_{1:T}=\left[
\be^{\rm EE}_1(x_1), \ldots, \be^{\rm EE}_{T_1}(x_{T_1}),
\bz_{T_1+1}^{(2:1)}(x_{T_1+1}), \ldots, \bz_{T}^{(2:1)}(x_{T})\right]^\top~\in ~ \R^{T \times b}.
\end{equation*}
This raw input tensor $\bx^\circ_{1:T}$ involves the following number of parameters
to be selected/fitted
\begin{equation*}
\varrho^{\rm input} = b \sum_{t=1}^{T_1} n_t + T_2 (2b + b(b+1)).
\end{equation*}
Having the input in this raw tensor structure $\bx^\circ_{1:T}\in \R^{T \times b}$
allows us to employ attention layers and Transformers to further process
this input data.

\subsubsection{Positional encoding}
Since attention layers do not have a natural notion of position and/or time in the tensor information, we complement the raw tensor $\bx^\circ_{1:T}$ by a  $b$-dimensional positional encoding; see Vaswani et al.~\cite{Vaswani} for positional encoding in the context of NLP.  However, we choose a simpler learned positional encoding compared to the sine-cosine encoding used by Vaswani et al.~\cite{Vaswani}, as this simpler version is sufficient for our purpose of learning on tabular data which do not have the natural ordering exploited in the original Transformer architecture. 

The positional encoding can be achieved by another embedding that considers the positions $t \in \{1,\ldots, T\}$ in a $b$-dimensional representation 
\begin{equation*}
\be^{\rm pos} : \{1,\ldots, T\} \to \R^b,
\qquad 
t  \mapsto  \be^{\rm pos}(t).
\end{equation*}
This positional encoding involves another $\varrho^{\rm position}=Tb$ parameters. We concatenate
this positional encoding with the raw input tensor providing us with the {\it input tensor}
\begin{eqnarray*}
\bx_{1:T}&=&
\begin{bmatrix}\bx^\circ_{1:T} & 
\begin{pmatrix}\be^{\rm pos}(1)^\top\\\vdots\\\be^{\rm pos}(T)^\top 
\end{pmatrix}\end{bmatrix}
\\&=&
\begin{bmatrix}
\be^{\rm EE}_1(x_1)& \cdots& \be^{\rm EE}_{T_1}(x_{T_1})&
\bz_{T_1+1}^{(2:1)}(x_{T_1+1})& \cdots& \bz_{T}^{(2:1)}(x_{T})\\
\be^{\rm pos}(1)&\cdots & \be^{\rm pos}(T_1)& \be^{\rm pos}(T_1+1)
&\cdots & \be^{\rm pos}(T) \end{bmatrix}^\top
~\in ~ \R^{T \times 2b}.
\end{eqnarray*}
We emphasize that all components of this input tensor $\bx_{1:T}$ originate from a specific input being either a covariate $x_t$ or the position $t$ of that covariate, i.e., it contains information about the individual instances. In the next subsection, we are going to add an additional variable, the CLS token, to this input tensor, but this additional variable does not originate from a covariate variable with a specific meaning.

\subsubsection{Classify (CLS) token}
\label{section Classify token}
\begin{figure}
    \centering
    \includegraphics[width=.8\linewidth]{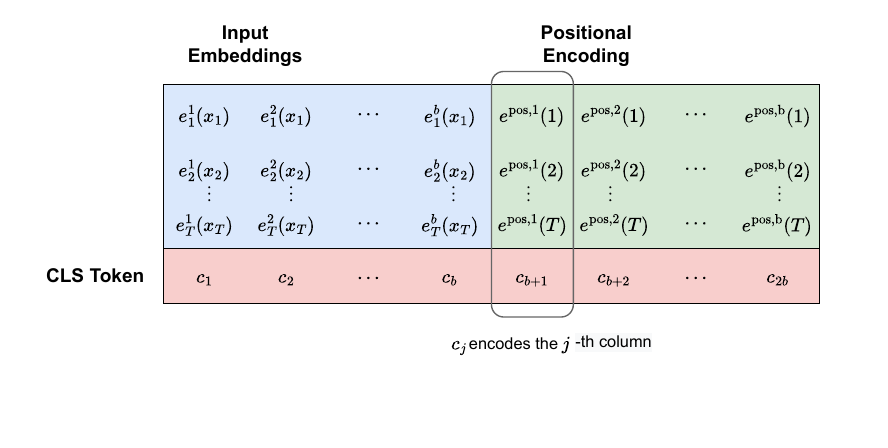}
    \vspace{-1cm}
    \caption{A diagram showing 
    the augmented input tensor $\bx^+_{1:T+1}$ consisting of
    input and positional encoding and the CLS token, moreover, it shows how the scalar components comprising the CLS token encode the columns of the input tensor $\bx_{1:T}$.
    }
    \label{fig:cls_diag}
\end{figure}

We extend the input tensor $\bx_{1:T}$ by an additional component, the CLS token. This is the step that differs from the classical Transformer proposal of Vaswani et al.~\cite{Vaswani}. The purpose of the CLS token is to encode every column $1\le j \le 2b$ of the input tensor $\bx_{1:T}\in \R^{T \times 2b}$ into a single variable.
This then gives us the {\it augmented input tensor}
\begin{eqnarray}\label{introduction CLS token}
\bx^+_{1:T+1}&=&
\begin{bmatrix}\bx_{1:T} \\ \bc^\top \end{bmatrix} 
\\&=&\nonumber
\begin{bmatrix}
\be^{\rm EE}_1(x_1)& \cdots& \be^{\rm EE}_{T_1}(x_{T_1})&
\bz_{T_1+1}^{(2:1)}(x_{T_1+1})& \cdots& \bz_{T}^{(2:1)}(x_{T})& \bc_1\\
\be^{\rm pos}(1)&\cdots & \be^{\rm pos}(T_1)& \be^{\rm pos}(T_1+1)
&\cdots & \be^{\rm pos}(T)& \bc_2 \end{bmatrix}^\top
~\in ~ \R^{(T+1) \times 2b}.
\end{eqnarray}
where $\bc=(\bc_1^\top,\bc_2^\top)^\top=(c_1,\ldots, c_{2b})^\top \in \R^{2b}$ denote the CLS tokens. 
Each of the scalars $c_j \in \R$ comprising the CLS tokens, $1\le j \le 2b$, will encode one column of the input tensor $\bx_{1:T} \in \R^{T\times 2b}$, i.e., it will provide a one-dimensional projection of the corresponding $j$-th $T$-dimensional vector to a scalar $c_j \in \R$; in text recognition models, one may imagine that the input tensor $\bx_{1:T}$ consists of $T$ embeddings of size $2b$, and the $j$-th dimension of each token embedding is summarized into a real-valued token $c_j$. For further processing, only the information contained in the CLS token $\bc$ will be forwarded to make predictions, as it reflects a compressed (encoded) version of the entire tensor information (after training of course). We illustrate the augmented input tensor $\bx^+_{1:T+1}$ in Figure \ref{fig:cls_diag}.

\subsection{Credibility Transformer layer}
\subsubsection{Transformer architecture} \label{Transformer archictecture}
The CLS token augmented input tensor $\bx^+_{1:T+1}\in \R^{(T+1) \times 2b}$ is first processed through a normalization layer (not indicated in our notation, but highlighted in Table \ref{architecture table}, below) before entering the Transformer architecture. Whereas several previous works in the actuarial literature have used batch-normalization \cite{Ioffe}, Transformer models usually use layer-normalization \cite{Ba}, which is what is utilized here.

We briefly describe the crucial modules of Transformers that are relevant for our Credibility Transformer architecture; for a more detailed description (and illustration) we refer to Vaswani et al.~\cite{Vaswani} and Richman--W\"uthrich \cite[Section 3.6]{RWJapan}.

Transformers are based on attention layers, and attention layers consist of queries $\bq_t$, keys $\bk_t$ and values $\bv_t$, $1\le t \le T+1$,
given by the FNN layer transformed inputs
\begin{eqnarray}\nonumber
  \bk_t &=& \phi\left(\bbb_K + W_K x^+_t\right)~\in~\R^{2b},\\
  \label{key, query, value}
  \bq_t &=& \phi\left(\bbb_Q +W_Q x^+_t\right)~\in~\R^{2b},\\
  \bv_t &=& \phi\left(\bbb_V +W_V x^+_t\right)~\in~\R^{2b},
            \nonumber
\end{eqnarray}
with weight matrices $W_K, W_Q, W_V \in \R^{2b\times 2b}$, biases $\bbb_K, \bbb_Q, \bbb_V \in \R^{2b}$, and where the activation function $\phi :\R \to \R$ is applied element-wise.
Since the weight matrices and biases are assumed to be $t$-independent,
we apply the same FNN transformation \eqref{key, query, value} in a time-distributed manner to all components $1\le t \le T+1$ of the augmented input tensor $\bx^+_{1:T+1}$. This provides us with tensors
\begin{eqnarray*}
K~=~K(\bx^+_{1:T+1}) &=& [ \bk_1, \ldots, \bk_{T+1}]^\top ~\in ~ \R^{(T+1)\times 2b},\\
Q~=~Q(\bx^+_{1:T+1}) &=& [ \bq_1, \ldots, \bq_{T+1}]^\top ~\in ~ \R^{(T+1)\times 2b},\\
V~=~V(\bx^+_{1:T+1}) &=& [ \bv_1, \ldots, \bv_{T+1}]^\top ~\in ~ \R^{(T+1)\times 2b}.
\end{eqnarray*}
The keys $K$ and queries $Q$ are used to construct the attention
weight matrix $A$ by applying the {\sf softmax} function to all rows (in a row-wise manner) in the following matrix
\begin{equation}\label{attention weight matrix}
  A =A(\bx^+_{1:T+1})= {\sf softmax}\left(\frac{Q K^\top}{\sqrt{2b}}  \right)~\in~ \R^{(T+1)\times (T+1)},\end{equation}
where the {\sf softmax} operation is defined for each row as
\begin{equation}\label{softmax definition}
  {\sf softmax}(\bz)_j = \frac{\exp(z_j)}{\sum_{k=1}^{T+1} \exp(z_k)}, \quad \text{for } j = 1, \ldots, T+1,
\end{equation}
with $\bz$ being a row of the matrix $QK^\top/\sqrt{2b}$.
  
Finally, the {\it attention head} of the Transformer is received by the matrix multiplication
\begin{equation}\label{attention head}
H=H(\bx^+_{1:T+1}) = A \, V ~\in~ \R^{(T+1) \times 2b}.
\end{equation}
This function encodes the augmented input tensor $\bx^+_{1:T+1}\in \R^{(T+1)\times 2b}$ in the attention head $H(\bx^+_{1:T+1}) \in \R^{(T+1) \times 2b}$ of
the same dimension. Essentially, this attention mechanism is a weighting
scheme reflected by the attention weights $A(\bx^+_{1:T+1})$
applied to the values $V(\bx^+_{1:T+1})=[ \bv_1, \ldots, \bv_{T+1}]^\top$. For our first application of the Transformer model, we only use one attention head, we modify this to a multi-head attention in Section \ref{Improving CT}, below.

A Transformer layer is then obtained by adding the attention head to the augmented input tensor resulting in a so-called skip-connection transformation
\begin{equation}\label{skip 1}
\bx^+_{1:T+1}~\mapsto ~ 
\bz^{\rm skip1}(\bx^+_{1:T+1})= \bx^+_{1:T+1}+H(\bx^+_{1:T+1})
  ~\in~ \R^{(T+1) \times 2b}.
\end{equation}
Typically, this transformed input is further processed by a time-distributed normalization layer
$\bz^{\rm norm}$, drop-out layers $\bz^{\rm drop}$ and post-processing 
time-distributed FNN layers $\bz^{\rm t-FNN}$ in combination with skip connections $\bz^{\rm skip}$; for full details we refer to Vaswani et al.~\cite{Vaswani}.
In our architecture, we process the
output $\bz^{\rm skip1}(\bx^+_{1:T+1})$
of \eqref{skip 1} further by applying the composed transformations
\begin{eqnarray}\label{Transformer}
\bz^{\rm trans}(\bx^+_{1:T+1})&=&\bz^{\rm skip1}(\bx^+_{1:T+1})
\\&+&
\left(\bz^{\rm norm2}\circ\bz^{\rm drop2}\circ\bz^{\rm t-FNN2}\circ\bz^{\rm drop1}\circ\bz^{\rm t-FNN1}\circ\bz^{\rm norm1}\right)
\left(\bz^{\rm skip1}(\bx^+_{1:T+1})\right).
\nonumber
\end{eqnarray}
Thus, we first normalize $\bz^{\rm norm1}$, then apply a time-distributed FNN layer followed by drop-out $\bz^{\rm drop1}\circ\bz^{\rm t-FNN1}$, and a second time-distributed FNN layer followed
by drop-out $\bz^{\rm drop2}\circ\bz^{\rm t-FNN2}$, 
and finally, we normalize again $\bz^{\rm norm2}$.
In a second skip-connection transformation we aggregate input and output of this transformation.
This Transformer architecture outputs a tensor shape $\R^{(T+1) \times 2b}$. We remark that the drop-out layers are only used during model fitting, to prevent the architecture from in-sample overfitting.

While the Transformer architecture is quite complex, we further remark that the model presented here is now quite standard in the machine learning literature. We provide a diagram for ease of understanding the various components of the Transformer in Figure \ref{fig:tf_diag} in the appendix.

\subsubsection{Credibility mechanism} \label{Credibility mechanism}
The remaining steps differ from Vaswani et al.~\cite{Vaswani}, and here, we take advantage of the integrated CLS tokens to complement this Transformer architecture $\bz^{\rm trans}(\cdot)$ by a credibility mechanism. For this we recall that the augmented input tensor $\bx^+_{1:T+1}$ considers
the (embedded) covariates $\bx=(x_t)_{t=1}^T$ and their positional encodings $(\be^{\rm pos}(t))_{t=1}^T$ in the first $T$ components, and it is complemented by the CLS tokens $\bc=\bx^+_{T+1} \in \R^{2b}$ in the $(T+1)$-st component of the augmented input tensor $\bx^+_{1:T+1}$, see \eqref{introduction CLS token}.

At this stage, i.e., before processing the CLS tokens using the Transformer, the CLS tokens do not carry any covariate specific information. Note that this is because before we have applied the Transformer layer, these CLS tokens $\bc=\bx^+_{T+1}$ have not interacted with the covariates; and
they are going to play the role of a global prior parameter in the following credibility consideration. For this purpose, we will extract the CLS token before the Transformer processing. After the Transformer  processing, these CLS tokens have interacted with the covariates through the attention mechanism \eqref{attention weight matrix}-\eqref{attention head}, and we are going to extract them a second time after this interaction, providing an embedded covariate summary. 

The first version of the CLS token is extracted from the value matrix $V=[ \bv_1, \ldots, \bv_{T+1}]^\top$. Importantly, to get this CLS token to be in the same embedding space as the outputs of the Transformer model, we need to process this first version of the CLS token with exactly the same operations (with the same weights) as in the Transformer \eqref{Transformer}, except that we do not need to apply the attention mechanism nor time-distribute the layers. This provides us with the following prior information
\begin{equation}\label{mean 1}
\bc^{\rm prior} = 
\left(\bz^{\rm norm2}\circ\bz^{\rm drop2}\circ\bz^{\rm FNN2}\circ\bz^{\rm drop1}\circ\bz^{\rm FNN1}\circ\bz^{\rm norm1}\right)
\left(\bv_{T+1}\right) ~\in~ \R^{2b}.
\end{equation}

Second, we extract the CLS token after processing through the Transformer \eqref{Transformer}, providing us with the tokenized information of the covariates $\bx$ and their positional embeddings
\begin{equation}\label{mean 2}
\bc^{\rm trans} =
\bz_{T+1}^{\rm trans}(\bx^+_{1:T+1})~\in~ \R^{2b},
\end{equation}
i.e., this is the $(T+1)$-st row of the Transformer output \eqref{Transformer}.
In this case, the CLS token has had the attention mechanism applied, and we emphasize that this version of the CLS token now contains a summary of the covariate information.

These two tokens \eqref{mean 1} and \eqref{mean 2} give us two different representations/predictors for the responses, the former representing only prior information, and the latter, prior information augmented by the covariates. We will use both of these tokens to make predictions from the Credibility Transformer by assigning weights to these representations.

This is done by selecting a fixed probability weight $\alpha \in (0,1)$ and then sampling during gradient descent training independent Bernoulli random variables $Z \sim {\rm Bernoulli}(\alpha)$ that encode which CLS token is forwarded to the rest of the network to make predictions, that is, we
forward the CLS token information according to
\begin{equation}\label{Credibility Transformer}
\bc^{\rm cred}=
Z\, \bc^{\rm trans} + \left(1-Z\right)\bc^{\rm prior}~\in~ \R^{2b}.
\end{equation}
Thus, in $\alpha\cdot 100\%$ of the gradient descent steps we forward the Transformer processed
CLS token $\bc^{\rm trans}$ which has interacted with the covariates, and in $(1-\alpha)\cdot 100\%$ of the gradient descent steps we forward the prior value CLS token $\bc^{\rm prior}$.
This can be seen as assigning a credibility of $\alpha$ to the Transformer token $\bc^{\rm trans}$ and the complementary credibility of $1-\alpha$ to the prior value $\bc^{\rm prior}$ of that
CLS token, to receive reasonable network parameters during gradient descent training. This mechanism \eqref{Credibility Transformer} is only applied during training, and for forecasting we set $Z \equiv 1$. The probability $\alpha \in (0,1)$ is treated as a hyperparameter that can be optimized by a grid search. We selected it bigger than $1/2$, because we would like to put more emphasis on the tokenized covariate information. 

\subsubsection{Rationale of credibility weighted CLS token} \label{rational}
We explain the rationale of the proposed credibility weighted CLS token $\bc^{\rm cred}$. There are two credibility mechanisms involved, a more obvious one that involves
the hyperparameter $\alpha \in (0,1)$, but, also, there is 
a second credibility mechanism involved in the CLS token
which learns an optimal credibility weight. This latter credibility mechanism arises from the prior information learned by the CLS token as a result of training the network with the hyperparameter $\alpha$. We now discuss these
two credibility mechanisms.

First, when training the Credibility Transformer, the CLS token $\bc^{\rm cred}$ will randomly be set to be either the token $\bc^{\rm trans}$, including covariate information, or the token $\bc^{\rm prior}$, encoding only prior information. In the latter case, the best prediction (in the sense of minimizing a deviance based loss function) that can be made without covariate information will be the portfolio mean. Thus, in  $(1-\alpha)\cdot 100\%$ of the training iterations of the Credibility Transformer, we will encourage the CLS token to incorporate prior information in the form of predicting the portfolio average. In the remaining iterations of the training, the CLS token will be trained on the covariate information to make more precise predictions than can be made using the portfolio average. 
In this sense, we can consider the CLS token $\bv_{T+1}$ before processing, i.e., as given in\eqref{mean 1},  representing the portfolio average in the embedding space of the rest of the tokens. 

\begin{figure}
    \centering
    \includegraphics[width=0.7\linewidth]{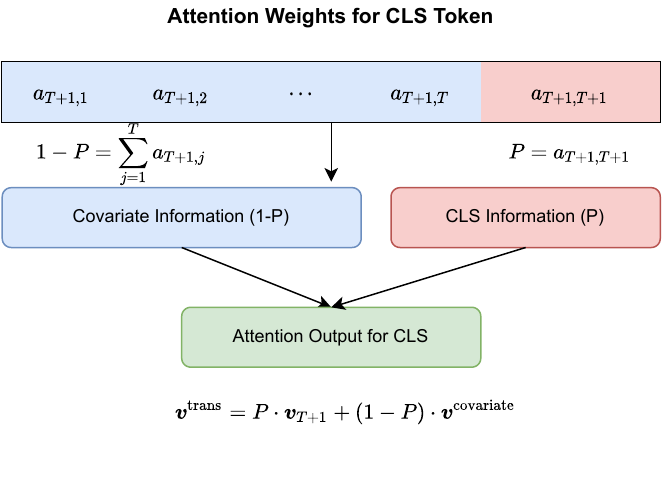}
    \vspace{-1cm}
    \caption{Diagram of how the attention mechanism can be viewed as a linear credibility formula.}
    \label{fig:ct_cred}
\end{figure}

We now dive deeper into the attention mechanism as it relates to the CLS token. Let $\ba_{T+1}=(a_{T+1,1},\ldots, a_{T+1,T+1})^\top \in \R^{T+1}$ denote the $(T+1)$-st row of the attention weight matrix $A$ defined in \eqref{attention weight matrix}. This row corresponds to the attention weights for the CLS token that forwards the extracted information. By using the {\sf softmax} function, we have normalization
\begin{equation*}
\sum_{j=1}^{T+1} a_{T+1,j} = 1.
\end{equation*}
We can interpret the last element of this vector, $a_{T+1,T+1}$, as the probability assigned to the CLS token itself, and we set
\begin{equation*}
P = a_{T+1,T+1}~\in ~ (0,1).
\end{equation*}
Consequently, the remaining probability $1-P$ is distributed among the covariate information
\begin{equation*}
\sum_{j=1}^{T} a_{T+1,j} =1-P.
\end{equation*}
This formulation allows us to interpret the attention mechanism in a way that is analogous to a linear credibility formula. Specifically, the attention output for the CLS token can be expressed as
\begin{equation}\label{attention output CLS}
\bv^{\rm trans} = P \, \bv_{T+1} + (1-P) \, \bv^{\rm covariate},
\end{equation}
where $\bv^{\rm trans}$ is the CLS embedding processed by the attention mechanism, in fact, this is the $(T+1)$-st row of the attention head $H$ given in
\eqref{attention head}, and the covariate information is a weighted sum of all values that consider the covariate information
\begin{equation}\label{covariate information}
 \bv^{\rm covariate} = \sum_{j=1}^{T}\, \frac{a_{T+1,j}}{1-P} \, \bv_{j}.
\end{equation}
This formulation demonstrates how the attention mechanism for the CLS token can be interpreted as a credibility weighted average of the CLS token's own information (representing collective experience) and the information from the covariates (representing individual experience), 
see Figure \ref{fig:ct_cred}. Essentially, this is a B\"uhlmann--Straub \cite{BS} type linear credibility formula, or a time-dynamic version thereof, with the credibility weights in this context learned and being input-dependent; this is similar to W\"uthrich \cite[Chapter 5]{Experience Rating}.

\subsection{Decoding of the tokenized information}
In the previous two sections we have encoded the covariate input $\bx$
in two steps to a tokenized variable $\bc^{\rm cred}$:
\begin{equation*}
\bx \qquad \mapsto \qquad\bx^+_{1:T+1}=\bx^+_{1:T+1}(\bx)\qquad \mapsto \qquad
\bc^{\rm cred}=\bc^{\rm cred}(\bx).
\end{equation*}
The final step is to decode this tokenized variable 
$\bc^{\rm cred}=\bc^{\rm cred}(\bx)$ so that it
is suitable to predict a response variable $Y$. This decoder
is problem-specific. Because our example below corresponds to
a one-dimensional single-task prediction problem, we use
a plain-vanilla FNN with a single output component to decode
the credibilitized token $\bc^{\rm cred}$.
This results in the decoder
\begin{equation*}
\bz^{(2:1)}: \R^{2b} \to \R, 
\qquad 
\bc^{\rm cred}(\bx) ~ \mapsto ~ \bz^{(2:1)}(\bc^{\rm cred}(\bx)),
\end{equation*}
where $\bz^{(2:1)}$ is a shallow FNN that maps from $\R^{2b}$
to $\R$. Since our responses $Y$ is a non-negative random variable,
we are going to model claim counts,
we add the exponential output activation, which all together results
in the {\it Credibility Transformer} (CT)
\begin{equation}\label{function Credibility Transformer}
\bx ~ \mapsto ~ \mu^{\rm CT}(\bx)=
\exp \left\{\bz^{(2:1)}(\bc^{\rm cred}(\bx))\right\}~>~0.
\end{equation}
The next section presents a real data example to exemplify
the use of the Credibility Transformer.

\section{Real data example}
\label{Real Data}
We benchmark the Credibility Transformer on
the commonly used French motor third party liability (MTPL) claims counts dataset of Dutang et al.~\cite{DutangCharpentier}.

\subsection{Description of data}
We perform the data pre-processing exactly as in W\"uthrich--Merz \cite{WM2023}, so that all our results can directly be benchmarked
by the results in that reference, and they are also
directly comparable to the ones in Brauer \cite[Table 2]{Brauer}.
For a detailed description of the French MTPL dataset 
we refer to
W\"uthrich--Merz \cite[Section 13.1]{WM2023}; and the
data cleaning and pre-processing is described in 
\cite[Listings 13.1 and 5.2]{WM2023}. We use the identical split into learning
and test datasets as in \cite{WM2023}.\footnote{The cleaned data
can be downloaded from \url{https://people.math.ethz.ch/~wueth/Lecture/freMTPL2freq.rda}, and for the partition into learning and test data
we use the random number generator under {\sf R} version 3.5.0.}
The learning dataset consists of 610,206 instances and the test dataset
of 67,801 instances; we refer to \cite[Section 5.2.4]{WM2023}.
This partition provides an aggregated time exposure of 322,392 calendar years
on the learning dataset, and 35,967 calendar years on the test dataset.
There are 23,738 claim events on the learning data, providing
an empirical frequency of $\widehat{\lambda}=7.35\%$, i.e., 
on average we observe an accident every $13.58=1/\widehat{\lambda}$ calendar years. This is a commonly observed expected frequency in European
MTPL insurance. On the test dataset we have 2,645 claim events providing
almost the same empirical frequency as on the learning dataset.

\subsection{Input tokenizer}
There are  $T_1=4$ categorical covariates with numbers of levels $n_t=6, 2, 11$ and $22$, 
and there are $T_2=5$ continuous covariates, providing
$T=T_1+T_2=9$ covariates. For embedding and tokenizing these input
covariates we select an embedding dimension of $b=5$.
This provides us with $\varrho^{\rm input} = 405$ weights
from the feature tokenizer giving us the raw input tensor $\bx_{1:9}^\circ
\in \R^{9\times 5}$. For the continuous input variable tokenizer we select
the linear activation function in the first FNN layers $\bz_t^{(1)}$ and the
 hyperbolic tangent activation function in the second FNN layers $\bz_t^{(2)}$.

\subsection{Description of the selected Credibility Transformer architecture}
Table \ref{architecture table} summarizes the Credibility Transformer architecture used.
\begin{table}[htb!]
\centering
{\small
\begin{center}
\begin{tabular}{|l||c|r|}
\hline
Module & Variable/layer & $\#$ Weights
\\\hline\hline
Feature tokenizer (raw input tensor) & $\bx_{1:9}^\circ$ & 405\\
Positional encoding & $\be^{\rm pos}_{1:9}$ & 45\\
CLS tokens & $\bc$ & 10\\
Time-distributed normalization layer & $\bz^{\rm norm}$ & 20\\
Credibility Transformer & $\bc^{\rm cred}$ & 1,073\\
FNN decoder & $\bz^{(2:1)}$& 193\\
\hline
\end{tabular}
\end{center}}
\caption{Credibility Transformer architecture used on the French MTPL dataset.}
\label{architecture table}
\end{table}

The total number of weights that need to be fitted is 1,746.
The Credibility Transformer uses 1,073 weights, these are 330 weights
from the time-distributed layers providing the keys $\bk_t$, queries $\bq_t$ and values $\bv_t$, 682
neurons from the two time-distributed FNN layers $\bz^{\rm t-FNN1}$
and $\bz^{\rm t-FNN2}$ having 32 and $2b=10$ neurons, respectively, and the remaining weights are from the normalization layers.
The decoder $\bz^{(2:1)}$  uses 193 weights coming from a first FNN layer 
having 16 neurons and the output layer, resulting
in the one-dimensional real-valued positive predictor \eqref{function Credibility Transformer}.
In all hidden layers we use the Gaussian error linear unit (GELU) 
activation function that takes the form $x \in \R \,\mapsto \, x\Phi(x)$ with the standard Gaussian distribution $\Phi$. The credibility parameter is set to
$\alpha=90\%$, this is an optimal value found by a grid 
search. For the drop-out layers we choose a drop-out rate of 1\%.
Both the credibility mechanism and the drop-out is only used during
network fitting, and for prediction we set $\alpha=1$ and the drop-out rate to zero.

\subsection{Gradient descent network fitting: 1st version}
\subsubsection{Plain-vanilla gradient descent fitting}
The Poisson deviance loss is used for claims counts model fitting. Minimizing the Poisson deviance
loss  is equivalent to maximum likelihood estimation (MLE) under a Poisson assumption. The average Poisson deviance loss is given by,
see W\"uthrich--Merz \cite[formula (5.28)]{WM2023},
\begin{equation}\label{Poisson deviance loss}
\frac{2}{n}\sum_{i=1}^n v_i \mu^{\rm CT}(\bx_i) - Y_i - Y_i
\log \left(\frac{v_i \mu^{\rm CT}(\bx_i)}{Y_i}\right)~\ge ~0,
\end{equation}
where $Y_i$ are the observed numbers of claims on policy $i$, 
$v_i>0$ is the time exposure of policy $i$, and $\mu^{\rm CT}(\bx_i)$
is the estimated expected claims frequency of that policy received
from the Credibility Transformer \eqref{function Credibility Transformer}.
The sum runs over all instances (insurance policies) $1\le i \le n$ that are included
in the learning data. Note that the Poisson deviance loss
is a strictly consistent loss function for mean estimation which is an important property that selected loss functions should possess for mean
estimation;
see Gneiting--Raftery \cite{GneitingRaftery} and Gneiting \cite{Gneiting}.

In our first fitting approach we use the {\tt nadam} version of stochastic
gradient descent (SGD) with its pre-selected parametrization implemented in the {\tt keras} package \cite{keras}. For SGD we use a batch size of $2^{10}=1,024$
instances, and since neural networks are prone to overfitting, we exploit an early stopping strategy by partitioning the learning data at random into
a training dataset and a validation dataset at a ratio of 9:1. We use the
standard {\tt callback} of {\tt keras} \cite{keras} using its pre-selected
parametrization to retrieve the learned weights with the smallest validation loss.
The optimal network weights are found after roughly 150 SGD epochs.

Neural network fitting involves several elements of randomness such as the (random) initialization of the SGD algorithm; see W\"uthrich--Merz \cite[Section 7.4.4]{WM2023}. To improve the robustness of the results
we always run 20 SGD fittings with different random initializations, and the reported results correspond to averaged
losses over the 20 different SGD fittings, in round brackets we
state the standard deviation of the losses over these 20 SGD fittings. 

Finally, in Richman--W\"uthrich \cite{RW2020} we have seen that one
can substantially improve predictive performance by ensembling over different
SGD runs. On average it takes 10 to 20 SGD fittings to 
get good ensemble predictors, see W\"uthrich--Merz \cite[Figure 7.19]{WM2023}. For this reason, in the last step of our procedure,  we
consider the ensemble predictor over the 20 different SGD runs.
From our results it can be verified that ensembling
significantly improves predictive models.

\subsubsection{Results of 1st fitting approach}
The results are given in Table \ref{results 1} (rows {\tt nadam}). These results
of the Credibility Transformer are  benchmarked with the ones
in W\"uthrich--Merz \cite[Tables 7.4-7.7 and 7.9]{WM2023} and Brauer
\cite[Tables 2 and 4]{Brauer}. We conclude that the Credibility Transformer clearly outperforms any of these other proposals out-of-sample, even the ensembled plain-vanilla FNN model is not much better than
a single run of the Credibility Transformer. 
By building the ensemble predictor of the Credibility Transformers
we receive a significantly better model providing an out-of-sample
average Poisson deviance loss of \bl{$23.717 \cdot 10^{-2}$},
see Table \ref{results 1}.
The best ensemble predictor in Brauer \cite[Table 4]{Brauer}
called CAFFT$_{\rm def}$ has an
average Poisson deviance loss of $23.726 \cdot 10^{-2}$.

\begin{table}[htb!]
\centering
{\small
\begin{center}
\begin{tabular}{|l||c|cccc|}
\hline
& \# &\multicolumn{2}{c}{In-sample\,\,}&\multicolumn{2}{c|}{Out-of-sample\,\,}\\
Model & Param.&\multicolumn{2}{c}{Poisson loss\,\,} & \multicolumn{2}{c|}{Poisson 
  loss\,\,} \\
\hline\hline
Poisson null  &1&25.213& &25.445&\\
Poisson GLM3 &50& 24.084 && 24.102&\\
Poisson plain-vanilla FNN &792& 23.728 & ($\pm$ 0.026)& 23.819&($\pm$ 0.017)\\
Ensemble Poisson plain-vanilla FNN &792& 23.691 && 23.783&\\\hline

Credibility Transformer: {\tt nadam}& 1,746& 23.648 &($\pm$ 0.071) & \bl{23.796} &($\pm$ 0.037)\\
Ensemble Credibility Transformer: {\tt nadam}& 1,746& 23.565 && \bl{23.717} &\\\hline
Credibility Transformer: {\tt NormFormer}& 1,746& 23.641 &($\pm$ 0.053) & \bl{23.788} &($\pm$ 0.040)\\
Ensemble Credibility Transformer: {\tt NormFormer}& 1,746& 23.562 && \bl{23.711} &\\
\hline
\end{tabular}
\end{center}}
\caption{Number of parameters, in-sample and out-of-sample Poisson deviance losses (units are in $10^{-2}$); benchmark models in the upper part of the table are taken
from \cite[Table 7.9]{WM2023}.}
\label{results 1}
\end{table}

\subsection{NormFormer gradient descent fitting: 2nd version}
Standard SGD optimizers such as {\tt nadam} with
a standard early stopping {\tt callback} do not work particularly well
on Transformers because of gradient magnitudes mismatches, 
typically
gradients on earlier layers are much bigger than gradients at later
layers; see Shleifer et al.~\cite{NormFormer}. Therefore, we
exploit the {\tt NormFormer} proposed by Shleifer et al.~\cite{NormFormer}
in our 2nd fitting attempt with the {\tt adam} version of SGD with
a learning rate of 0.002 and $\beta_2=0.98$; reducing this parameter of {\tt adam} has been suggested, e.g., in Zhai et al.~\cite{zhai2023sigmoid}.
The results are given in the lower part of Table \ref{results 1}.
We observe a slight improvement in prediction accuracy,
the Credibility Transformer ensemble having a smaller out-of-sample
average Poisson deviance loss of \bl{$23.711 \cdot 10^{-2}$}.

\begin{figure}[htbp]
  \centering
  \includegraphics[width=.84\textwidth]{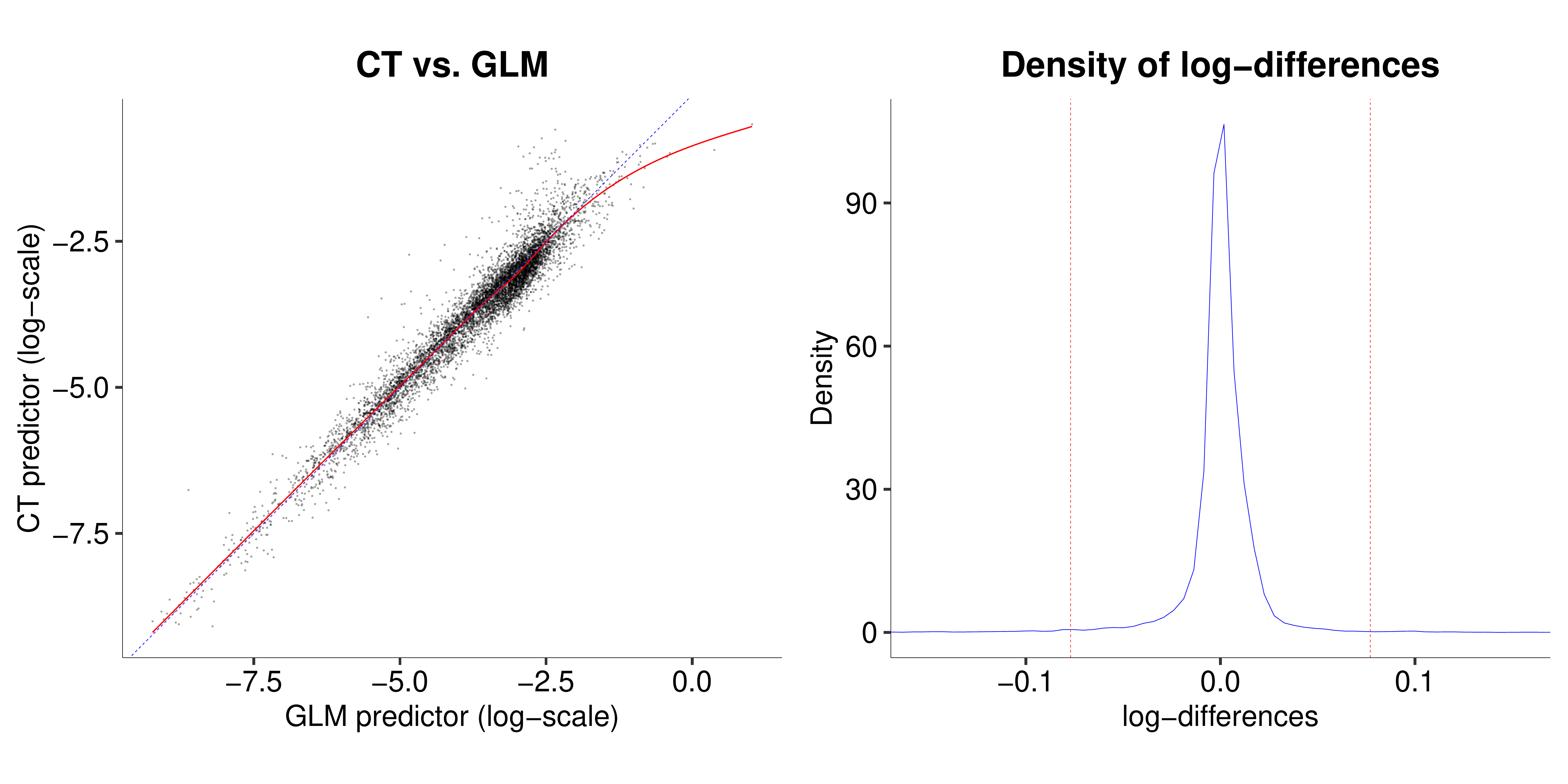}
  \caption{(lhs) Scatterplot of predictions from one run of the Credibility Transformer (CT)
 vs.~GLM predictions; (rhs) density
of log-differences of the two predictors, the vertical dotted
lines correspond to 2 empirical standard deviations.}
  \label{scatterplot 1}
\end{figure}

Figure \ref{scatterplot 1} (lhs) gives a scatterplot that compares
one run of the Credibility Transformer predictions ({\tt NormFormer} SGD fitting) against the
GLM predictions (out-of-sample). The red line gives a GAM smoother. On average the two predictors are rather similar, except
in the tails. However, the black cloud of predictors
has individual points that substantially differ from the darkgray
diagonal line, indicating that there are bigger differences on an
individual insurance policy scale. This is verified by the density
plot of the individual log-differences of the predictors on the right-hand side of Figure
\ref{scatterplot 1}.

\begin{figure}[htb!]
\begin{center}
\begin{minipage}[t]{0.42\textwidth}
\begin{center}
\includegraphics[width=\textwidth]{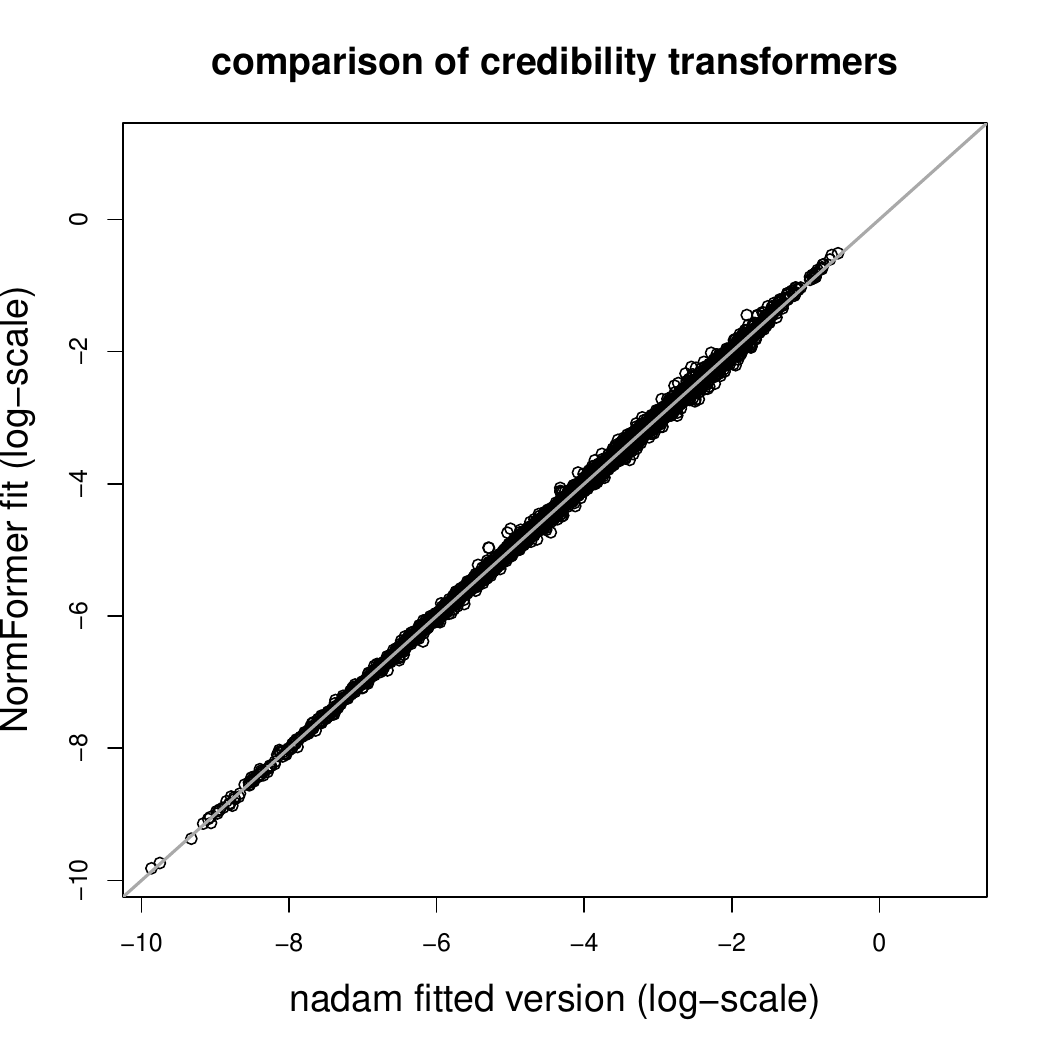}
\end{center}
\end{minipage}
\end{center}
\caption{Scatterplot of ensemble Credibility Transformer predictions:
{\tt NormFormer} SGD fitting against {\tt nadam} fitting.}
\label{scatterplot 3}
\end{figure}

Figure \ref{scatterplot 3} compares the two ensemble Credibility
Transformer predictions obtained from the {\tt NormFormer} SGD
fitting procedure and the {\tt nadam} fitting procedure, the former
slightly outperforming the latter in terms of out-of-sample loss. The figure shows that the individual 
predictions lie fairly much on the diagonal line saying that there
are not any bigger individual differences. This supports robustness
of the fitting procedure.

We emphasize that the credibility mechanism \eqref{Credibility Transformer}
is only applied during SGD training. For prediction, we turn this 
credibility mechanism off by setting $\alpha=100\%$, resulting in 
weights $Z\equiv 1$ in \eqref{Credibility Transformer}, thus, only the observation based
CLS token $\bc^{\rm trans}$ is considered for prediction. We may ask whether the network
has also learned any structure for the prior 
CLS token $\bc^{\rm prior}$. This can be checked by setting
$\alpha=0\%$, resulting in weights $Z\equiv 0$
in \eqref{Credibility Transformer}. Using the resulting
predictor we receive a constant prediction of $\widehat{\lambda}=7.35\%$, i.e., the homogeneous model without considering any covariate information.
Of course, this is expected by the design of the network architecture.
However, it is still worth to verify this property to ensure that the network architecture
works as expected. If we set
$Z\equiv 90\%$ to the same rate as used for network fitting we
receive a bigger out-of-sample loss of $23.727 \cdot 10^{-2}$ not
supporting credibility robustification of the predictor, once properly trained.

\section{Improving the Credibility Transformer}
\label{Improving CT}
We now work to further improve the Credibility Transformer. The following modifications can be categorized into three main areas. First, we augment the Transformer with a multi-head attention and utilize a deep version of the architecture. Second, we use insights from the LLM literature to more flexibly handle inputs to the model by using gated layers. Third, we modify the inputs to the network by using a differentiable continuous covariate embedding layer based on a novel idea from Gorishny et al.~\cite{Gorishniy}. Through applying these changes, we achieve a very strong out-of-sample performance of the Credibility Transformer. We start by discussing each of these changes and then return to demonstrate the performance of the improved architecture. 

\subsection{Multi-head attention and deep Transformer architecture}
\subsubsection{Multi-head attention}
In Section \ref{Transformer archictecture}, we have described a Transformer model with a single attention head, see \eqref{attention head}. An important extension of this simpler model is {\it multi-head attention} (MHA), which applies the attention mechanism  \eqref{attention weight matrix}-\eqref{attention head} multiple independent copies of the query, key and value projections of the input data. This allows the model to attend to information from different learned representations. After learning these multiple representations, these are projected back to the original dimension of the model, which is $2b$ in our case.

Formally, we chose the number of attention heads $M \in \N$. The MHA is defined as 
\begin{equation}\label{multi-head attention breakdown}
H^{\text{MHA}}(\bx^+_{1:T+1}) = \text{Concat}(H_1, \ldots, H_M) W^O ~\in~ \R^{(T+1) \times 2b},
\end{equation}
where we concatenate multiple attention heads 
to a tensor
\begin{equation*}
\text{Concat}(H_1, \ldots, H_M) = [H_1, H_2, \ldots, H_M] ~\in~ \R^{(T+1) \times (M d)},
\end{equation*}
with attention heads $H_m \in \R^{(T+1) \times d}$
of dimension $d \in \N$, for $1 \le m \le M$, and output projection matrix $W^O \in \R^{(M d) \times 2b}$. Typically, one
chooses $M d = 2b$, in other words, the total dimension of the model is chosen for $Md$, and it is proportionally allocated to each attention head.

For each head, $1 \le m \le M$, we use the same formulation as above for the attention operation, 
with attention head specific keys, queries and values
\begin{eqnarray*}
H_m &=& A_m \, V_m ~\in~ \R^{(T+1) \times d}, 
\\
A_m &=& {\sf softmax}\left(\frac{Q_m K_m^\top}{\sqrt{d}}\right) ~\in~ \R^{(T+1) \times (T+1)}, 
\\
K_m &=& \phi\left(\bbb^{(m)}_K + \bx^+_{1:T+1} W_K^{(m)}\right) ~\in~ \R^{(T+1) \times d},
\\
Q_m &=& \phi\left(\bbb^{(m)}_Q + \bx^+_{1:T+1} W_Q^{(m)} \right) ~\in~ \R^{(T+1) \times d}, 
\\
V_m &=& \phi\left(\bbb^{(m)}_V + \bx^+_{1:T+1} W_V^{(m)} \right) ~\in~ \R^{(T+1) \times d}. 
\end{eqnarray*}

At this point, we have described a standard implementation of a MHA. To improve the stability of fitting this model we again follow Shleifer et al.~\cite{NormFormer} by adding a multiplicative scaling coefficient to each head before it enters into \eqref{multi-head attention breakdown}, i.e., we update this to 
\begin{equation}\label{multi-head attention normformer}
H^{\text{MHA}}(\bx^+_{1:T+1}) = \text{Concat}(\alpha_1 H_1, \ldots, \alpha_M H_M) W^O ~\in~ \R^{(T+1) \times 2b},
\end{equation}
where $\alpha_m \in (0,1]$, for $1 \le m \le M$. During training, we apply dropout to $\alpha_m$ to improve the out-of-sample performance of the network.
After this MHA modification, the rest of the Transformer layer used here is the same as in Sections \ref{Transformer archictecture} and \ref{Credibility mechanism}. 

\subsubsection{Deep Credibility Transformer}
We use multiple Transformer layers composed together to create a deep version of the model applied above, assuming that we have Transformer layers $1 \le \ell \le L$, for $L\in \N$. To compose these layers, while retaining the ability to use the credibility mechanism \ref{Credibility Transformer}, we need to modify the inputs to the Transformer layers. The first layer of these is exactly the same as presented above, producing outputs $\bz^{\rm trans, 1}(\bx^+_{1:T+1})$, including the CLS token $\bc^{\rm trans, 1}$, which could be stripped out if needed at this point, and $\bc^{\rm prior, 1}$, where we have added an upper index $^1$ to show that this is the first Transformer layer with its outputs.

For the second and subsequent layers, $2 \le \ell \le L$, we input $\bz^{\rm trans, \ell-1}$ and $\bc^{\rm prior, \ell-1}$ recursively into layer specific versions of \eqref{Transformer} 
\begin{eqnarray}\label{Transformer_deep}
\bz^{\rm trans, \ell}(\bx^+_{1:T+1}) &=& \bz^{\rm skip1,\ell}(\bz^{\rm trans, \ell-1}(\bx^+_{1:T+1})) \nonumber \\
&&+~ \left(\bz^{\rm norm2, \ell}\circ\bz^{\rm drop2, \ell}\circ\bz^{\rm t-FNN2, \ell}\circ\bz^{\rm drop1, \ell}\circ\bz^{\rm t-FNN1, \ell}\circ\bz^{\rm norm1, \ell}\right) \nonumber  \\
&&\qquad \circ ~ \left(\bz^{\rm skip1, \ell}(\bz^{\rm trans, \ell-1}(\bx^+_{1:T+1}))\right)  ~\in~ \R^{(T+1) \times 2b},
\end{eqnarray}
and \eqref{mean 1}\footnote{For notational convenience, we have not included the head-specific key, query and value projections of $\bc^{\rm prior, \ell-1}$ to derive $\bc^{\rm prior, \ell}$, nonetheless, these are also performed.}
\begin{equation}\label{mean 1_deep}
\bc^{\rm prior, \ell} = 
\left(\bz^{\rm norm2, \ell}\circ\bz^{\rm drop2, \ell}\circ\bz^{\rm FNN2, \ell}\circ\bz^{\rm drop1, \ell}\circ\bz^{\rm FNN1, \ell}\circ\bz^{\rm norm1, \ell}\right)
\left(\bc^{\rm prior, \ell-1}\right) ~\in~ \R^{2b}.
\end{equation}
Finally, the credibility weighting between the outputs is performed
\begin{equation}\label{Deep Credibility Transformer}
\bc^{\rm cred, L}=
Z\, \bc^{\rm trans, L} + \left(1-Z\right)\bc^{\rm prior, L}~\in~ \R^{2b}.
\end{equation}
The same rationale of linear credibility as given in Section \ref{rational} applies here too; in each attention head (a projection of) the prior information in the CLS token is reweighted with the (projected) covariate information being processed there.
For an illustration, see Figure \ref{fig:deep_tf_diag} in the appendix.

\subsection{Gated layers}
The model presented to this point uses a standard FNN to process the outputs of the attention mechanism. Modern LLMs usually rely on a more advanced mechanism incorporating a gating principle for the FNN; this has been introduced to the LLM literature by Dauphin et al.~\cite{Dauphin17a}, with the concept of a Gated Linear Unit (GLU). The main difference between the usual FNNs and a GLU is that not all of the inputs to a FNN have equal importance, and, indeed, some of the less important inputs should be down-weighted or even removed from the computations performed by the network. A GLU accomplishes this by adding a second FNN with a sigmoid activation function - which produces outputs between 0 and 1 - to the usual FNN, and then multiplies these 
by an (element-wise) Hadamard product $\odot$, i.e., 
\begin{equation} \label{GLU}
    \bz^{\rm GLU}(\bx) = \bz^{\rm FNN_{sigmoid}}(\bx)\odot \bz^{\rm FNN_{linear}}(\bx) ,
\end{equation}
where the GLU layer $\bz^{\rm GLU}$ is the Hadamard
product of a standard affine projection $\bz^{\rm FNN_{linear}}$
and a FNN layer with sigmoid activation
$\bz^{\rm FNN_{sigmoid}}$. Beyond this simple formulation of the GLU, modern LLMs usually follow Shazeer \cite{Shazeer2020}, who replaces the sigmoid activation in \eqref{GLU} with a different activation function. Especially popular is to replace the sigmoid function with a so-called sigmoid linear unit (SiLU) activation function, due to Elfwing et al.~\cite{elfwing2018sigmoid}, defined as
\begin{equation*}
    \text{\rm SiLU}(x) = x \odot \sigma(x).
\end{equation*}
The SiLU-GLU layer, abbreviated to SwiGLU for
Swish GLU, see Shazeer \cite{Shazeer2020}, is defined as
\begin{equation*} 
    \bz^{\rm SwiGLU}(\bx) = \bz^{\rm FNN_{linear}}(\bx) \odot \bz^{\rm FNN_{SiLU}}(\bx).
\end{equation*}
In our implementation, we replace the first FNNs in equations \eqref{Transformer_deep} and \eqref{mean 1_deep} with these SwiGLU layers. This modification allows one for more complex interactions between features and improves the model's ability to down-weight less important components of the input data.

\subsection{Improving the continuous covariate embedding}
The final major modification to the Credibility Transformer is to improve the continuous covariate embedding \eqref{numerical embeddings layers} by replacing the simple two-layer FNN with a more advanced approach. We start with the Piecewise Linear Encoding (PLE) of Gorishny et al.~\cite{gorishniy2022embeddings}, which encodes continuous covariates using a numerical approach that, in principle, is similar to an extension of one-hot encoding to continuous variables. The main idea of PLE is to split the range of each continuous covariate into bins, and then encode the covariate value based on which bin it falls into. This provides a more expressive representation compared to the original scalar values. 

Formally, for the $t$-th continuous covariate of the $T_2$ continuous covariates available in the dataset, we partition its range of values into $B_t \in \N$ disjoint intervals - which are called {\it bins} - ${\mathfrak B}_t^j = [b_{t}^{j-1}, b_t^j)$, for $1\le j \le B_{t}$, and with $b_{t}^{j-1}< b_t^j$. The PLE encoding
of covariate value $x_t$ is then defined by the vector
\begin{equation*}
{\rm PLE}_t(x_t) = \left(e^{\rm PLE}_1, \ldots, e^{\rm PLE}_{B_t}\right)^\top ~\in ~\R^{B_t},
\end{equation*}
where for $1\le j \le B_{t}$ 
\begin{equation*}
e^{\rm PLE}_j ~=~ e^{\rm PLE}_j(x_t) ~=~ \frac{x_t - b_t^{j-1}}{b_t^j - b_t^{j-1}}\,
\mathds{1}_{\{x_t \in {\mathfrak B}_t^j\}}
+\mathds{1}_{\{x_t \geq b_t^j\}}.
\end{equation*}
This encoding gives a $B_t$-dimensional vector ${\rm PLE}_t(x_t)$, which has components 1 as long as $x_t \ge b_t^{j}$, for $x_t \in {\mathfrak B}_t^j$ we interpolate linearly, and for $x_t < b_t^{j-1}$ the
components of ${\rm PLE}_t(x_t)$ are zero.
This encoding has several desirable properties, among which, we note that it provides a more expressive yet lossless representation of the original scalar values that preserves the ordinal nature of continuous covariates. In the original proposal of Gorishny et al.~\cite{gorishniy2022embeddings}, the bin boundaries are determined either using quantiles of the observed values of the covariate components $x_t$ or they are derived from a decision tree. Differently to these approaches, we introduce a differentiable version of PLE that allows for end-to-end training of the bin boundaries. This approach enables the model to learn optimal bin placements for each continuous covariate during the training process. The key idea is to parameterize the bin boundaries indirectly through trainable weights that define the length of each bin. The bin boundaries $(b_t^j)_{j=0}^{B_t}$ are computed as
\begin{equation*}
b_t^j = s_t + \sum_{k=0}^j \delta_t^k,
\end{equation*}
where $s_t\in \R$ is a given (fixed) starting value, and $\delta_t^k>0$ are learnable bin lengths.
To ensure numerical stability and enforce non-negative bin widths, we parameterize the bin lengths using their logarithms
\begin{equation*}
\delta_t^k = \exp(\log(\delta_t^k)).
\end{equation*}
The logged bin lengths $\log(\delta_t^k)$ are initialized randomly or from pre-computed initial bins using quantiles. Finally, to handle potential numerical issues and to ensure a minimum bin length, we introduce a threshold $\epsilon>0$
\begin{equation*}
\delta_t^k ~\leftarrow ~\delta_t^k \,\mathds{1}_{\{\delta_t^k \ge \epsilon\}}
\end{equation*}
Thus, if the size of the learned bin is less than $\epsilon$, we collapse the bin into the previous one.
This differentiable PLE allows the model to adapt the bin boundaries during training, which, taken together with the next part, allows more effective feature representations to be learned.

\begin{figure}[htbp]
      \centering
      \includegraphics[width=0.6\textwidth]{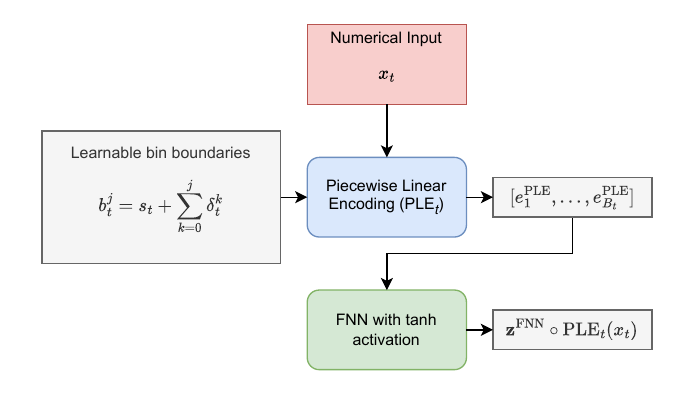}
      \caption{Visualization of differentiable Piecewise Linear Encoding (PLE) for a numerical feature.}
      \label{ple_visualization}
  \end{figure}

To use the PLE in the Transformer model, following Gorishny et al.~\cite{gorishniy2022embeddings}, we add a trainable FNN, 
$\bz_t^{\rm FNN}$, with a hyperbolic tangent activation after the PLE to produce the final feature embedding
\begin{equation*}
x_t~\mapsto~
\be^{\rm NE}_t(x_t) = \left(\bz_t^{\rm FNN} \circ {\rm PLE}_t\right)(x_t) ~\in~ \R^b, 
\end{equation*}
where $\be_t^{\rm NE}$ is the numerical embedding (NE) from the PLE layer. We expect that the PLE followed by a 
hyperbolic tangent layer is more effective than a simple two-layer FNN for continuous covariate embeddings since, unlike a FNN which applies a global transformation to its inputs, this numerical embedding captures local patterns within different intervals of the covariate's range. This is because the FNN with a hyperbolic tangent activation expands the information contained in the PLE into a fixed-size embedding $\be_t^{\rm NE} \in \R^b$, which varies linearly within bins but non-linearly across different bins.
A diagram of the differentiable PLE  is shown in Figure \ref{ple_visualization},

\subsection{Other modifications}
Here, we briefly mention some of the other less complex modifications made to the Credibility Transformer. Two of these are inspired by Holzm\"uller et al.~\cite{holzmuller2024better}. First, continuous inputs to the network are scaled by subtracting the median of the data and then dividing by the inter-quartile range; this provides inputs that are more robust to outliers. Second, we allow for learned feature selection by scaling each embedding by a constant in the range $(0,1]$ before these enter the Transformer. We initialize all FNN layers that are followed by the GeLU activation function using the scheme of He et al.~\cite{he2015delving} and use the {\tt adamW} optimizer of Loshchilov--Hutter \cite{loshchilov2017decoupled}, with weight decay set to $0.02$. Finally, we set the $\beta_2=0.95$ in the optimizer, which is a best practice to stabilize optimization with Transformer architectures that use large batches and versions of the {\tt adam} optimizer, see Zhai et al.~\cite{zhai2023sigmoid}.

\subsection{Results of the improved deep Credibility Transformer}
The improved version of the Credibility Transformer was fit to the same split of the French MTPL dataset as above. Compared to the details discussed in Section \ref{Real Data}, we follow the same approach, except that we fit on batches of size $2^{12} = 4096$ instances. We set the number of attention heads in each layer $M=2$ and use three transformer layers, i.e., we use a deep Transformer architecture. Moreover, we set $b=40$, in other words, we use a model that is eight times wider than the one used above. Since this results in a very large number of parameters to optimize - approximately 320,000 -  it becomes infeasible to fit these models without utilizing Graphics Processing Units (GPUs); the approach taken was to utilize a cloud computing service to fit these models. Each training run of the model takes about 7 minutes\footnote{This was done on an L4 GPU on the Google Colab service.}, i.e, fitting such a large and complex model is entirely feasible using GPUs. We show the in-sample and out-of-sample Poisson deviance losses for each credibility parameter in Table \ref{results_table 2}, where we evaluate the model's performance for different values of the credibility parameter $\alpha$, ranging from $90\%$ to $100\%$. Recall that setting the credibility parameter to $\alpha = 100\%$ corresponds to the plain-vanilla Transformer training approach without
using the credibility mechanism for training.

\begin{table}[htb!]
\centering
{\small
\begin{center}
\begin{tabular}{|l||cc|cc|}
\hline
& \multicolumn{2}{c|}{In-sample} & \multicolumn{2}{c|}{Out-of-sample} \\
Model & \multicolumn{2}{c|}{Poisson loss} & \multicolumn{2}{c|}{Poisson loss} \\
\hline\hline
Ensemble Poisson plain-vanilla FNN  &23.691 && 23.783&\\
Ensemble Credibility Transformer (best-performing) & 23.562 & & 23.711 & \\
\hline
Improved Credibility Transformer with $\alpha = 90\%$ & 23.533 & ($\pm$ 0.058) & 23.670 & ($\pm$ 0.031) \\
Ensemble Credibility Transformer ($\alpha = 90\%$) & 23.454 & & 23.587 & \\
Improved Credibility Transformer with $\alpha = 95\%$ & 23.557 & ($\pm$ 0.058) & 23.676 & ($\pm$ 0.027) \\
Ensemble Credibility Transformer ($\alpha = 95\%$) & 23.465 & & 23.593 & \\
Improved Credibility Transformer with $\alpha =98\%$ & \bl{23.544} & \bl{($\pm$ 0.042)} & \bl{23.670} & \bl{($\pm$ 0.032)} \\
Ensemble Credibility Transformer ($\alpha =98\%$) & \bl{23.460} & & \bl{23.577} & \\
Improved Credibility Transformer with $\alpha = 100\%$ & 23.535 & ($\pm$ 0.051) & 23.689 & ($\pm$ 0.044) \\
Ensemble Credibility Transformer ($\alpha = 100\%)$ & 23.447 & & 23.607 & \\
\hline
\end{tabular}
\end{center}}
\caption{In-sample and out-of-sample Poisson deviance losses 
(units are $10^{-2}$) for each credibility parameter applied to the improved
deep Credibility Transformer of Section \ref{Improving CT};
ensembling results are shown for each credibility parameter.}
\label{results_table 2}
\end{table}

From Table \ref{results_table 2} we observe that the 
different credibility parameters $\alpha
\in \{90\%, 95\%, 98\%, 100\%\}$ lead to similar in-sample Poisson deviance losses, and they are relatively stable, with minor fluctuations. Specifically, the in-sample losses are all very close to $23.544 \cdot 10^{-2}$. The standard deviations are also comparable, indicating consistent performance across different runs.

When examining the out-of-sample Poisson deviance losses, we notice a slightly different pattern. The losses slightly decrease as the credibility parameter increases from $90\%$ to $98\%$, reaching the lowest value at $\alpha = 98\%$ with an average loss of \bl{$23.670 \cdot 10^{-2}$} (standard deviation $\pm 0.032$). Setting $\alpha = 98\%$ yields the best out-of-sample performance among the individual models, suggesting that incorporating the credibility mechanism enhances the model's generalization ability. Comparing the models with and without the credibility mechanism, we find that the models with $\alpha < 100\%$ generally outperform the plain-vanilla Transformer model ($\alpha = 100\%$) in out-of-sample predictions. This indicates that the credibility mechanism effectively leverages prior information, improving the predictive accuracy on unseen data. 

The benefits of ensemble modeling through averaging are evident from the results, i.e., even with a state-of-the-art Transformer model, significant gains can be made through ensembling. This is a somewhat surprising result, since the usual practice with large Transformer-based models is usually to use only one model. The ensemble models consistently achieve lower Poisson deviance losses compared to their individual counterparts. For example, the ensembling architecture with $\alpha = 98\%$ attains the lowest out-of-sample loss of \bl{$23.577 \cdot 10^{-2}$}, outperforming both the individual models and the plain-vanilla Transformer ensemble approach. 

Moreover, in comparison to the original Credibility Transformer presented earlier, the improved version shows substantial performance gains. The best-performing model with $\alpha = 98\%$ achieves an out-of-sample loss of \bl{$23.577 \cdot 10^{-2}$}, which is a significant improvement over the original model's performance of $23.711 \cdot 10^{-2}$. We observe that the model's performance appears to be somewhat sensitive to the credibility parameter, with optimal performance achieved at $\alpha = 98\%$. This suggests that while the credibility mechanism is beneficial, it is helpful to fine-tune this parameter for optimal results. 

We also observe that the out-of-sample standard deviation of the Poisson deviance loss is somewhat higher in the plain-vanilla training approach, i.e., when the credibility parameter is set to $100\%$. Thus, we see that the proposed credibility approach helps to stabilize the training process.

While the improved model's complexity necessitates the use of GPUs, the performance gains likely justify the increased computational requirements for many practical applications in insurance pricing and risk assessment. However, practitioners should carefully consider the trade-offs between model complexity, performance, and interpretability in their specific contexts.

In summary, the improved deep Credibility Transformer with a credibility parameter $\alpha$ slightly less than $100\%$, i.e., incorporating the credibility regularization, achieves superior predictive performance. We thus conclude that even in a state-of-the-art Transformer model, as presented here, incorporating the credibility mechanism can improve the out-of-sample performance of the model.

\begin{figure}[htbp]
  \centering
  \includegraphics[width=.84\textwidth]{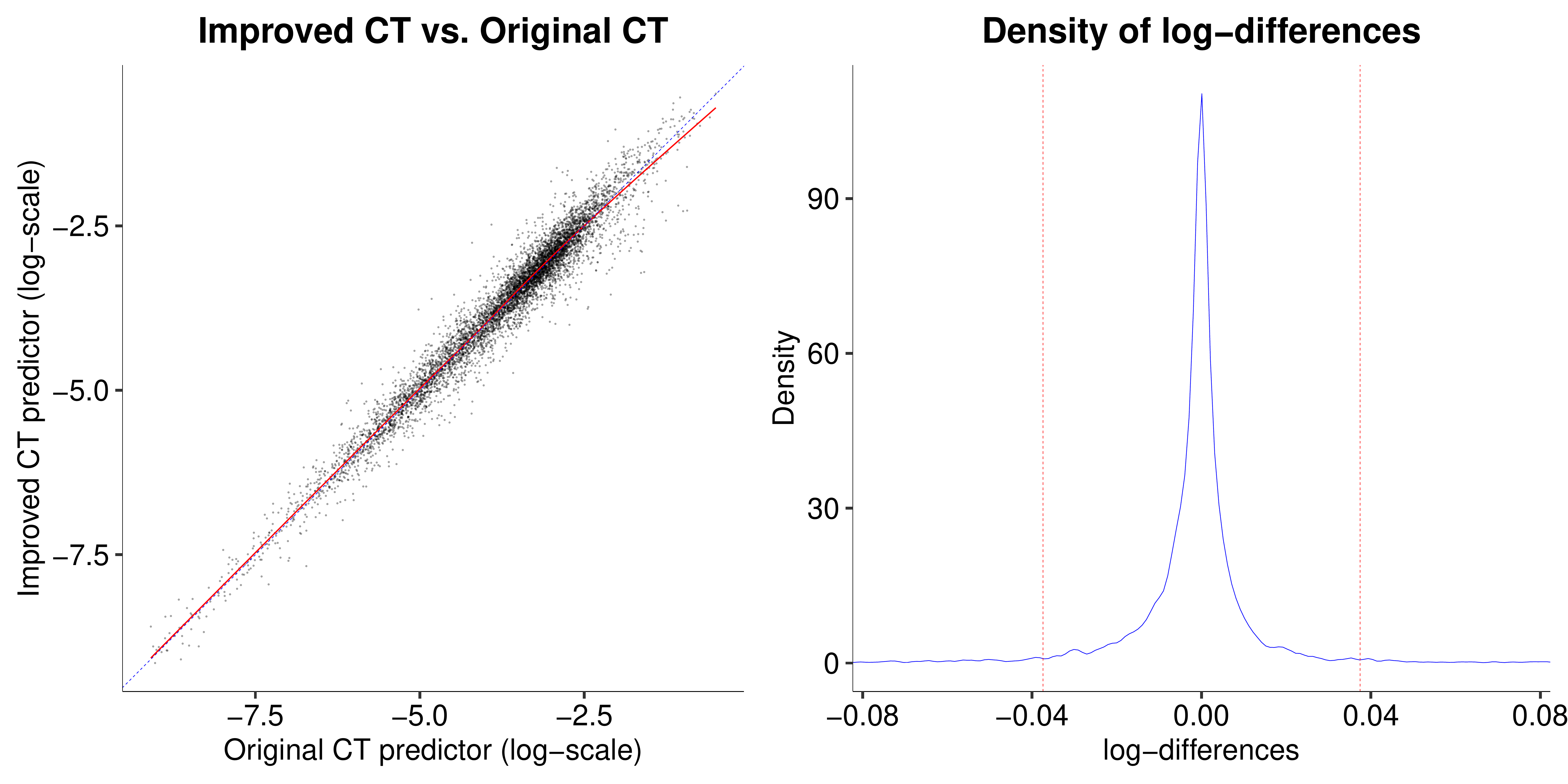}
  \caption{(lhs) Scatterplot of predictions from one run of the 
  (original) Credibility Transformer
 vs. one run of the improved deep Credibility Transformer; (rhs) density of log-differences of the two predictors, the vertical dotted
lines correspond to 2 empirical standard deviations.
}
  \label{scatterplot 2}
\end{figure}

Figure \ref{scatterplot 2} (lhs) gives a scatterplot that compares
the original Credibility Transformer predictions ({\tt NormFormer} fitting) against the
improved deep Credibility Transformer predictions (out-of-sample). Similar to the above, the red line gives a local GAM smoother. We see that, on average the two predictors are rather similar, in this case, even in the tails. We further see there are bigger differences on an
individual insurance policy scale; nonetheless, the density
plot of the individual log-differences of the predictors on the right-hand side of Figure
\ref{scatterplot 2} shows that these differences are smaller than the differences between the GLM and Credibility Transformer shown in Figure \ref{scatterplot 1}.

\section{Exploring the Credibility Transformer predictions}
\label{Exploring CT}
The Credibility Transformer architecture provides rich information about how the model predictions are formed, in particular, we refer to Section \ref{rational} for a discussion of how the attention operation that updates the CLS token can be seen as a linear credibility formula. We focus on the single-head variations of Section \ref{Model architecture section} for simplicity; with some effort, the same analysis can be produced for the improved deep Credibility Transformer, after aggregating the attention outputs across heads and layers. Here, we select one trained Credibility Transformer model to explore the workings of this model and note that we have selected a model that performs similarly to the out-of-sample Poisson deviance loss reported in Table \ref{results 1} for the {\tt NormFormer} variation. 

\begin{figure}[htbp]
  \centering
  \includegraphics[width=0.49\textwidth]{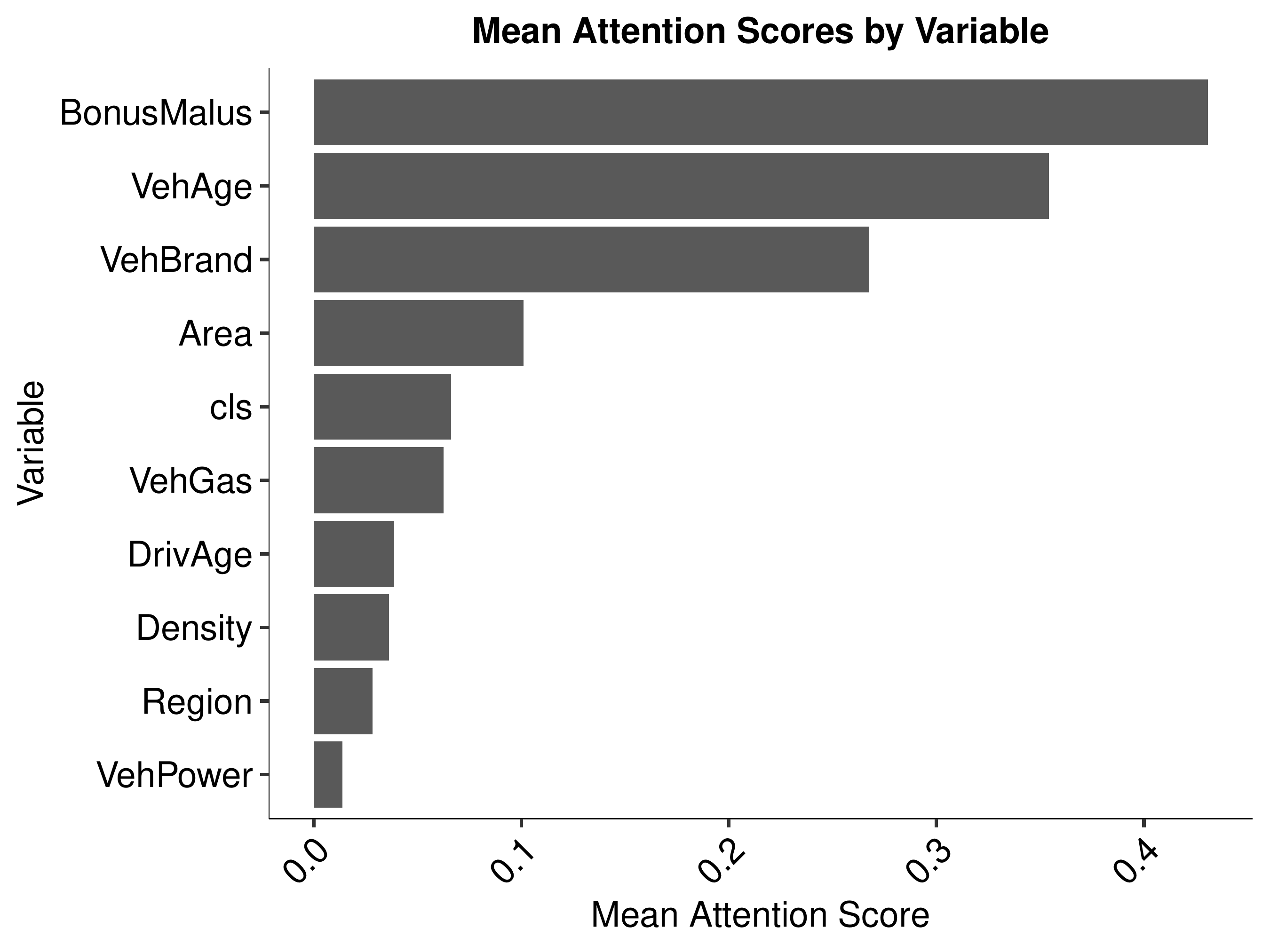}
  \includegraphics[width=0.49\textwidth]{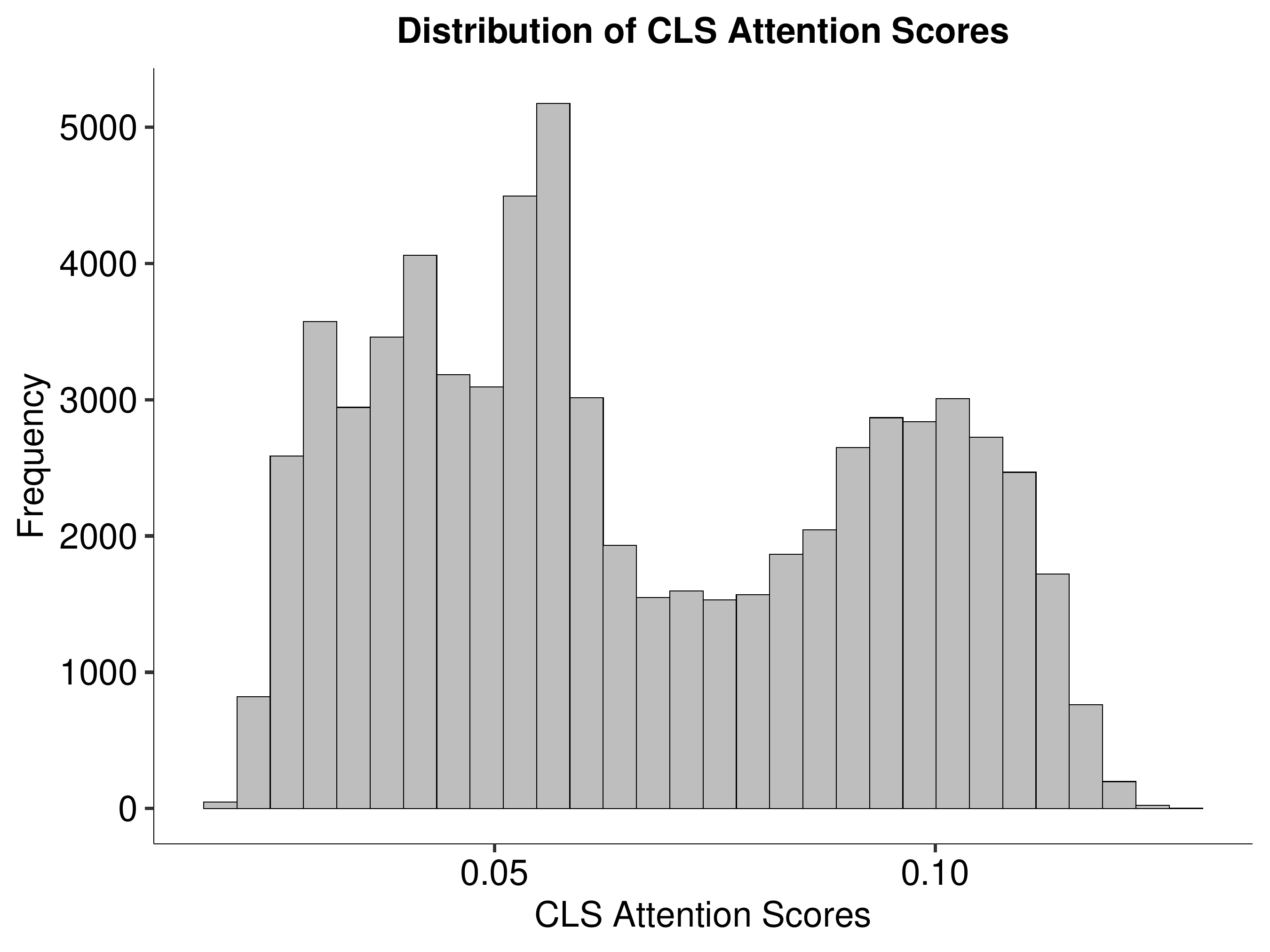}
  \caption{(lhs) Average values of the attention outputs $(a_{T+1,j})_{j = 1}^{T+1}$ evaluated on the test dataset for a Credibility Transformer model; (rhs) histogram of the values taken by $P = a_{T+1,T+1}$  evaluated on the test dataset for a Credibility Transformer model across all individual policies.}
  \label{cls attn hist}
\end{figure}

We start by examine the mean values of the attention outputs for the CLS token, $(a_{T+1,j})_{j = 1}^{T+1}$. These mean values are produced by taking the average over the test dataset used for assessing the model's performance. In Figure \ref{cls attn hist} (lhs), we show that the variable to which the model attends the most is the {\tt BonusMalus} score, followed by {\tt VehAge}, {\tt VehBrand} and {\tt Area} code. The CLS token, which we recall is calibrated to produce the portfolio average frequency, has an attention probability of about 6.5\% on average, i.e., with reference to equation \eqref{attention output CLS}, $P = a_{T+1,T+1}$ is comparably low, implying that the complement probability $1-P$, which is the weight given to the covariates, is relatively high. This is exactly as we would expect for a large portfolio of MTPL insurance policies. In Figure \ref{cls attn hist} (rhs), we show a histogram of the attention outputs for $a_{T+1,T+1}$, which is the credibility factor $P$. The histogram across the test dataset which takes values between zero and approximately $12.5\%$, says that even in the most extreme cases, only a comparably small weight is placed on the portfolio average experience.

\begin{figure}[htbp]
  \centering
  \includegraphics[width=.9\textwidth]{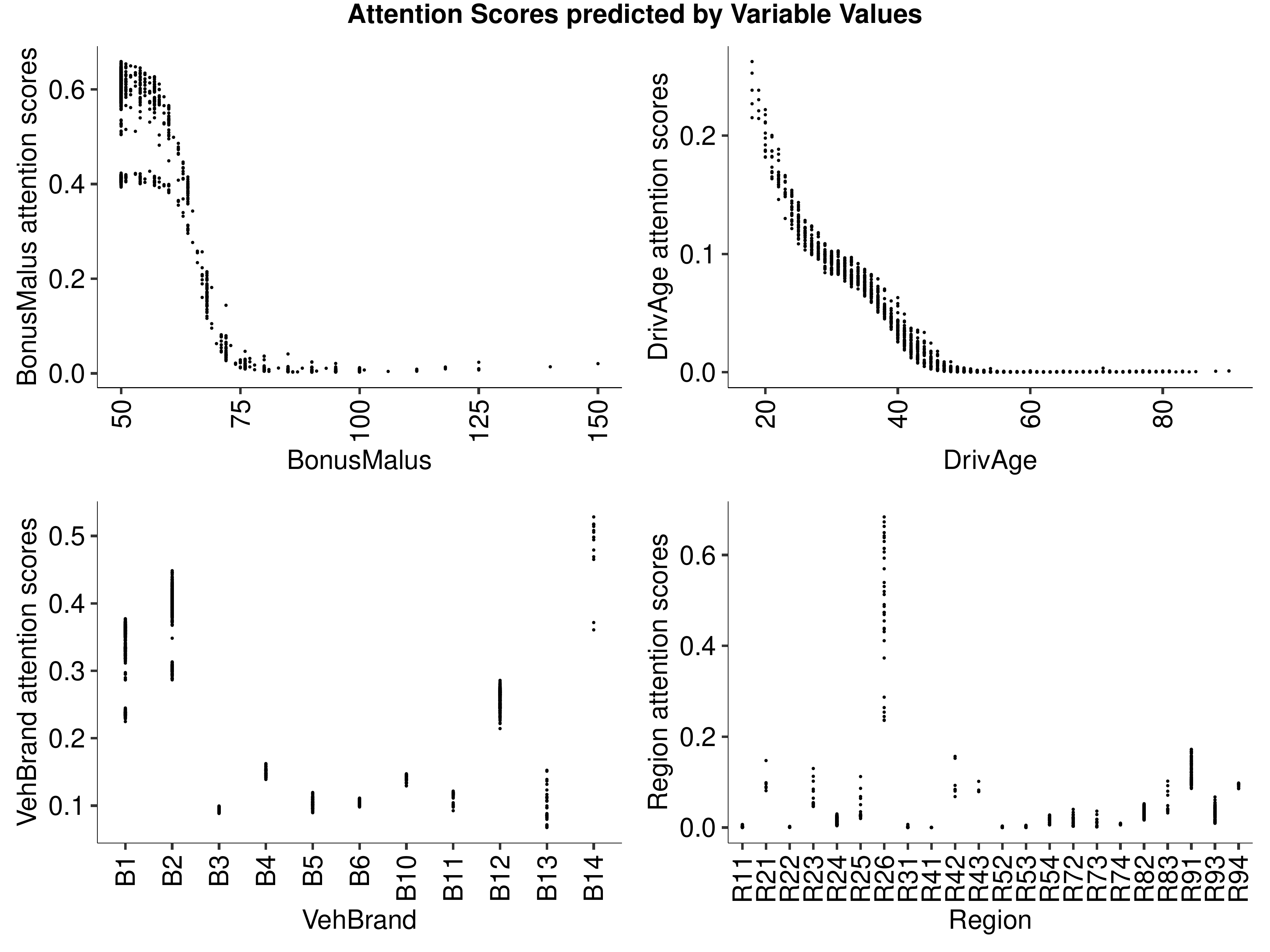}
  \caption{Relationships learned by the Credibility Transformer model between the covariates and the attention scores $a_{T+1,j}$ for the {\tt BonusMalus}, {\tt DrivAge}, {\tt VehBrand} and {\tt Region} covariates.}
  \label{attn versus covariates}
\end{figure}

In Figure \ref{attn versus covariates}, we investigate the relationships learned by the model between the covariates and the attention scores; this is done for two continuous and two categorical variables, namely, {\tt BonusMalus}, {\tt DrivAge}, {\tt VehBrand} and {\tt Region} (see the next paragraph for a description of how these variables were chosen). The figures show interesting relationships which indicate that the model attends strongly to the information contained within the embedding of these variables, for certain values of the covariate. For example, low values of the (correlated) {\tt BonusMalus} and {\tt DrivAge} covariates receive strong attention scores. Likewise, some values of the {\tt VehBrand} and {\tt Region} covariates receive very strong attention scores. These relationships can be investigated further by trying to understand the context in which variations may occur, e.g., in the top left panel of Figure \ref{attn versus covariates} we can see that, at low values of the {\tt BonusMalus} covariate, two different attention patterns occur for some segments of the data. In Figure \ref{attn versus covariates, bm}, we reproduce the top left panel of Figure \ref{attn versus covariates}, but color the points according to the {\tt Density} covariate, revealing that drivers with a low {\tt BonusMalus} score in high density regions have lower attention scores for the {\tt BonusMalus} covariate, than drivers in medium and low {\tt Density} areas, in other words, these variables interact quite strongly.

\begin{figure}[htbp]
  \centering
  \includegraphics[width=0.55\textwidth]{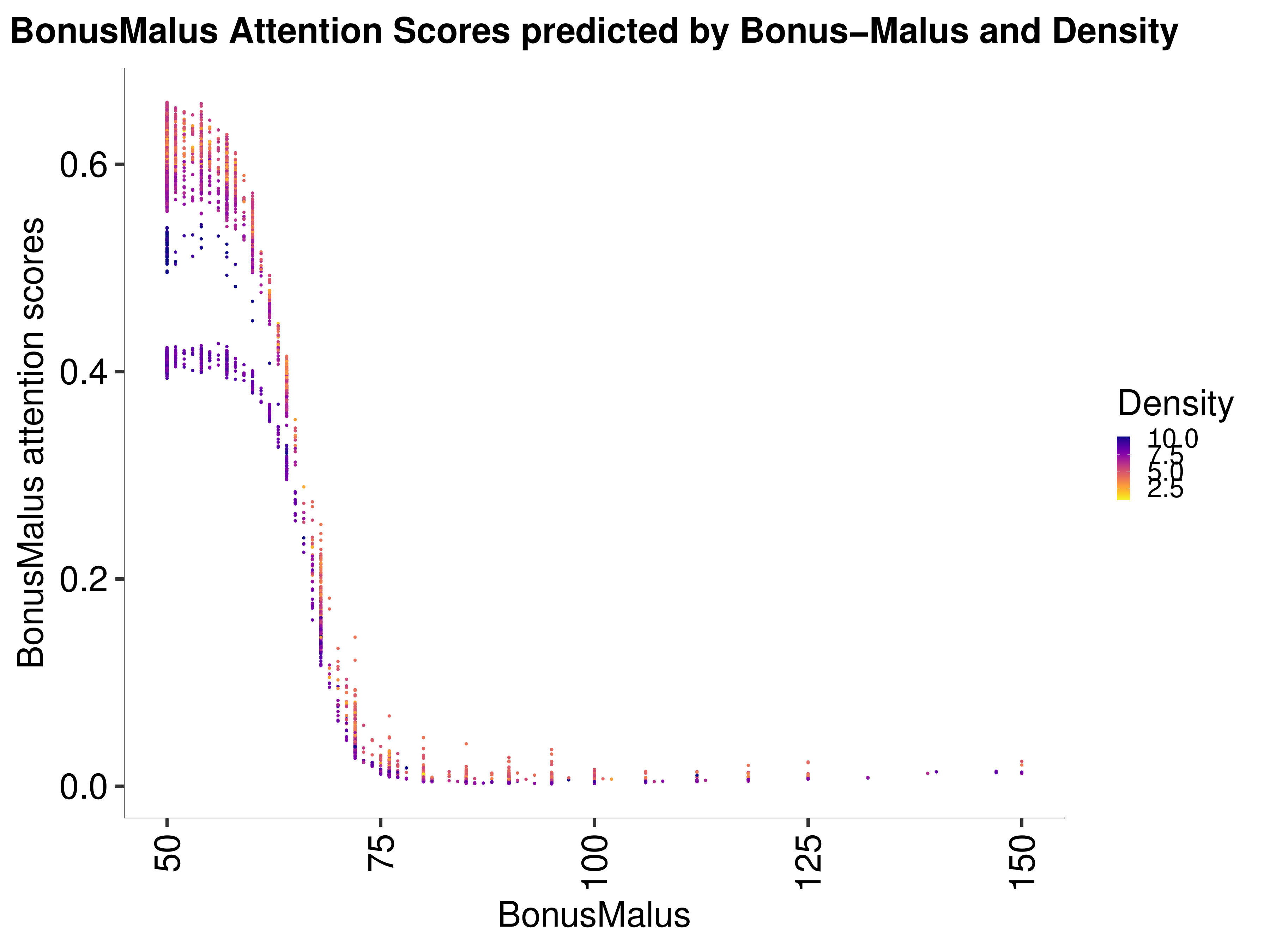}
  \caption{Relationships learned by the Credibility Transformer model between the {\tt BonusMalus} covariate and the attention scores, points colored by the value of the {\tt Density} covariate.
}
  \label{attn versus covariates, bm}
\end{figure}

\begin{figure}[htbp]
  \centering
  \includegraphics[width=.8\textwidth]{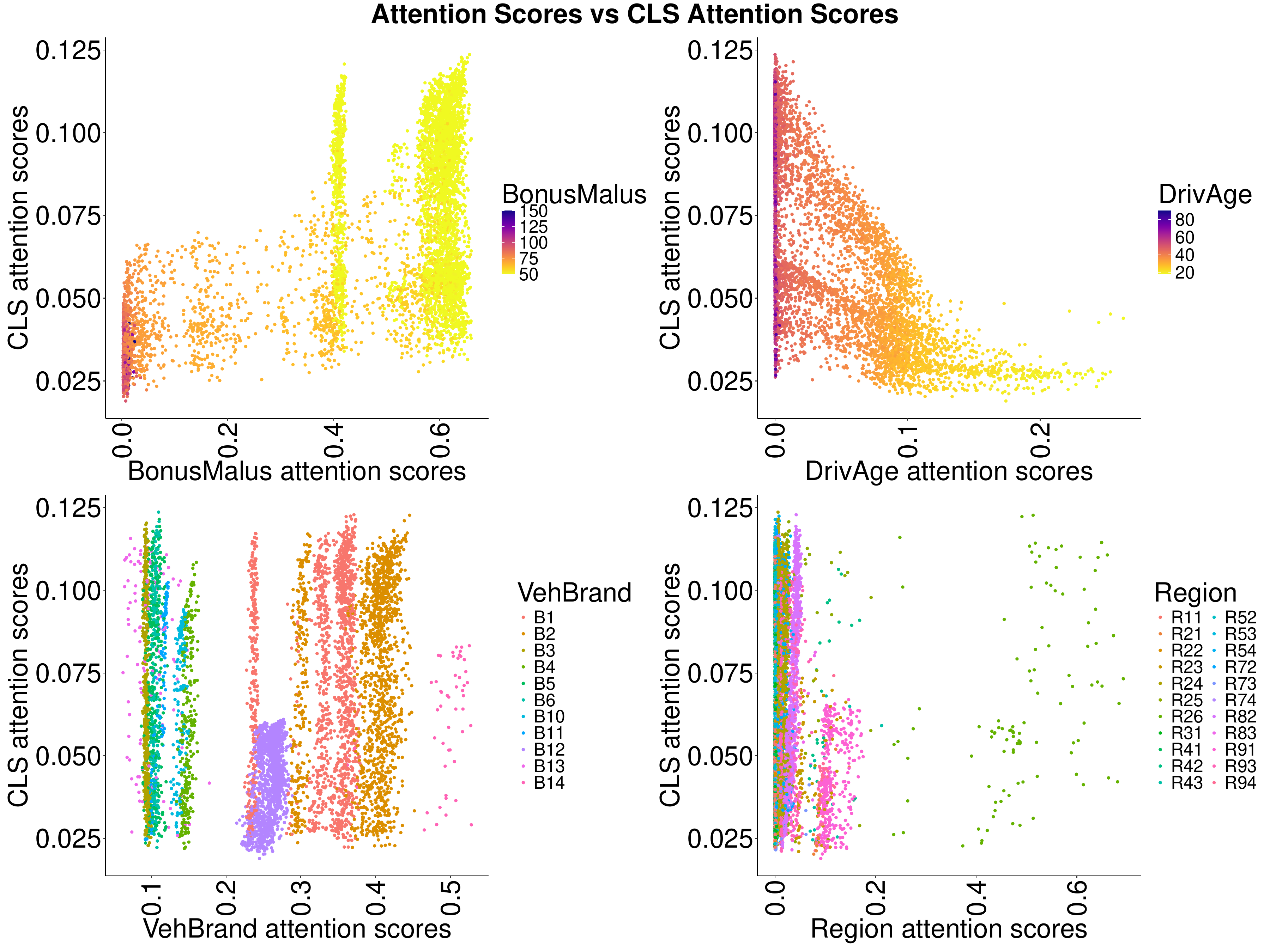}
  \caption{Relationships between the attention scores for selected covariates and the CLS token shown for the {\tt BonusMalus}, {\tt DrivAge}, {\tt VehBrand} and {\tt Region} covariates; points are colored according to the value taken by the covariate under consideration.}
  \label{cls attn versus covariate attn}
\end{figure}

\begin{figure}[htbp]
  \centering
  \includegraphics[width=0.55\textwidth]{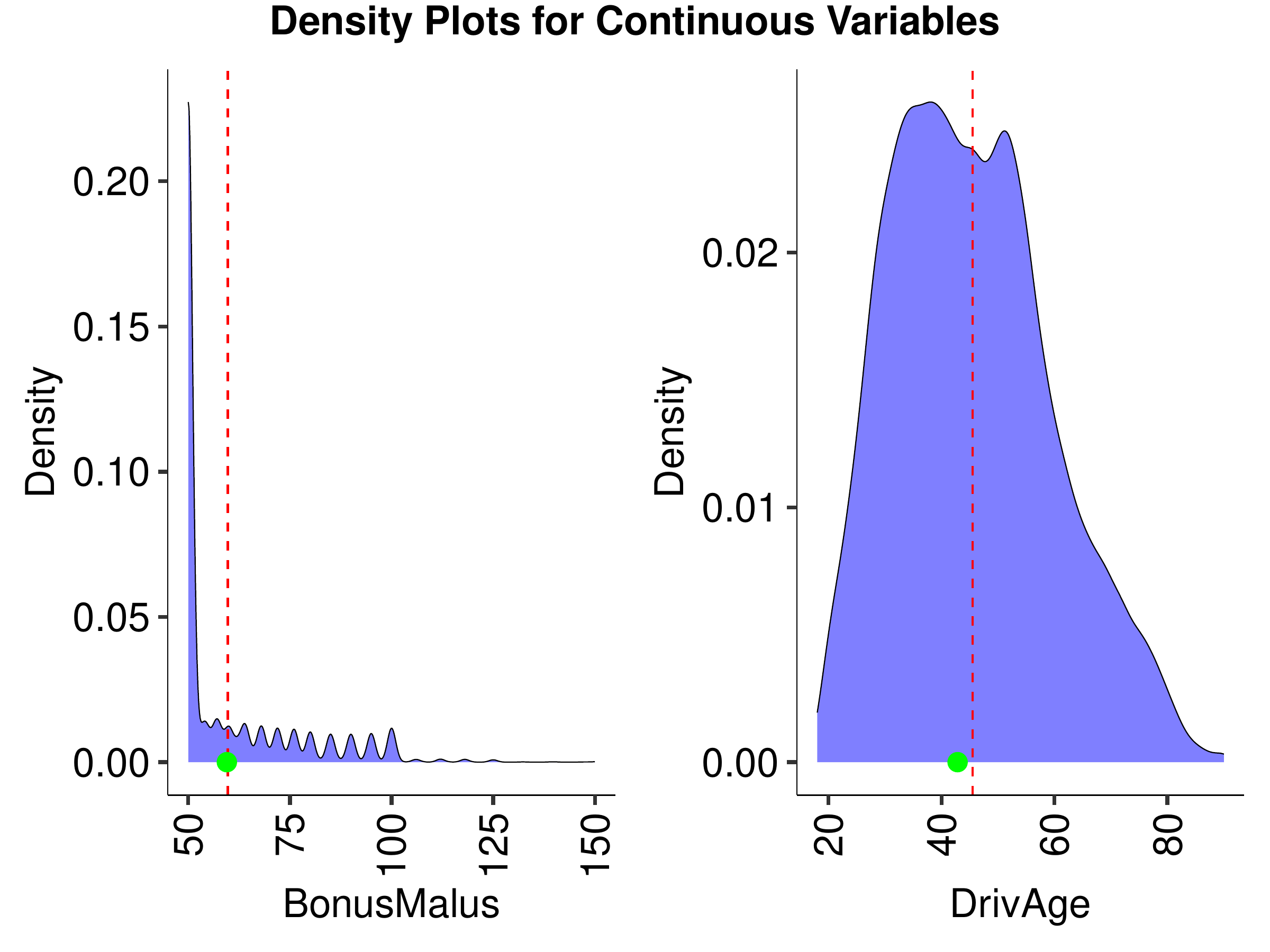}
  \caption{Density plot of the values of the {\tt BonusMalus} (lhs) and {\tt DrivAge} (rhs) covariates; dotted red line indicates the average value of the covariate and the green dot represents the average value of the covariate producing predictions close to the predicted portfolio mean. }
  \label{density and mean}
\end{figure}

Finally, in Figure \ref{cls attn versus covariate attn}, we show how the attention scores for the CLS token vary with the attention scores given to the other covariates, for the {\tt BonusMalus}, {\tt DrivAge}, {\tt VehBrand} and {\tt Region} covariates. These covariates were selected as the top four covariates maximizing the variable importance scores of a Random Forest model fit to predict the CLS token attention scores based on the attention scores for the other covariates. The analysis shows a very strong relationship between the covariate and the CLS attention scores, i.e., for how the credibility given to the portfolio experience varies with the values of the other covariates. In particular, we can see that the highest credibility is given to the portfolio experience when {\tt BonusMalus} scores are low, when {\tt DrivAge} is middle-aged, when {\tt VehBrand} is not {\tt B12}, and in several of the {\tt Regions} in the dataset. In summary, this analysis leads us to expect that - at least to some extent - the Credibility Transformer  produces frequency predictions close to the portfolio average for these, and similar, values of the covariates. We test this insight in Figure \ref{density and mean}, which shows density plots of the values of the {\tt BonusMalus} and {\tt DrivAge} covariates, as well as the average value of the covariate and the average value of the covariate producing predictions close to the predicted portfolio mean. It can be seen that for low values of the {\tt BonusMalus} covariate, and for middle-aged drivers, the Credibility Transformer produces predictions close to the portfolio mean. As just mentioned, for these ``average" policyholders, we give higher credibility to the CLS token.

\section{Conclusions}
\label{Conclusions}
In this paper, we introduced and developed the Credibility Transformer, a novel approach to using Transformers for actuarial modeling in a non-life pricing context, which integrates traditional credibility theory with state-of-the-art deep learning techniques for tabular data processing using Transformers. We have demonstrated that the Credibility Transformer consistently outperforms standard Transformer models in out-of-sample prediction as measured by the Poisson deviance loss. This improvement is significant and robust across multiple runs, illustrating the effectiveness of our approach in enhancing predictive accuracy for insurance pricing. Moreover, our results support the use of ensembling techniques even when applying complex state-of-the-art Transformer models since it was shown that ensemble models consistently achieved lower Poisson deviance losses compared to individual models.

Building on the initial Credibility Transformer that was introduced using a single-head attention within a shallow model, the enhanced version of the Credibility Transformer incorporates several architectural enhancements, including multi-head attention, gated layers, and improved numerical embeddings of continuous covariates. These modifications collectively contributed to the model's superior performance. While the improved model's complexity necessitates the use of GPUs to train the model, the performance gains appear to justify the increased computational requirements. Moreover, the ability to train these complex models in a reasonable time-frame (approximately 7 minutes per run) on cloud-based GPUs makes them feasible for real-world deployment.

The consistent performance across different credibility parameters and the reduced out-of-sample standard deviation (compared to plain-vanilla Transformers) indicate that the approach introduced here enhances the stability and reliability of predictions. Thus, we can conclude that this work demonstrates a successful integration of traditional actuarial concepts (credibility theory) with state-of-the-art deep learning techniques.

Finally, an analysis of the rich information provided by the model shows that, in line with expectations on a personal lines motor insurance portfolio, the trained Credibility Transformer gives only minor weight to the CLS token representing the portfolio's average experience and that, for example, interactions can be identified using the attention scores of the model.

Future research directions could include developing methods to dynamically adjust the credibility parameter during training or even on a per-instance basis.

\medskip

{\bf Acknowledgement.}
Parts of this research were carried out while Mario W\"uthrich was a KAW guest professor at Stockholm University, and while he
was hosted at Ewha Womans University, Seoul.

\medskip

{\small 
\renewcommand{\baselinestretch}{.51}
}

\newpage
\appendix
\begin{figure}
    \centering
    \includegraphics[width=0.5\linewidth]{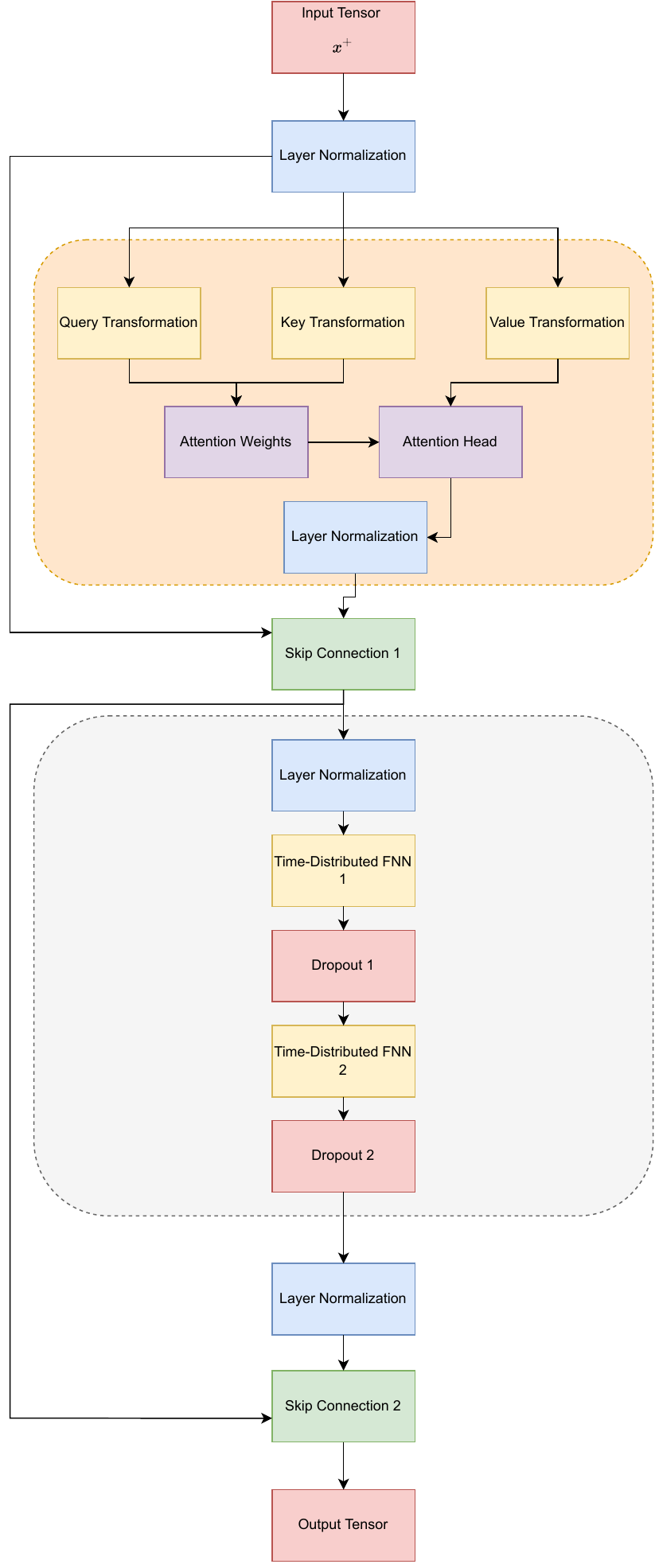}
    \caption{Diagram of the Transformer architecture \eqref{skip 1}-\eqref{Transformer}.}
    \label{fig:tf_diag}
\end{figure}

\begin{figure}
    \centering
    \includegraphics[width=0.5\linewidth]{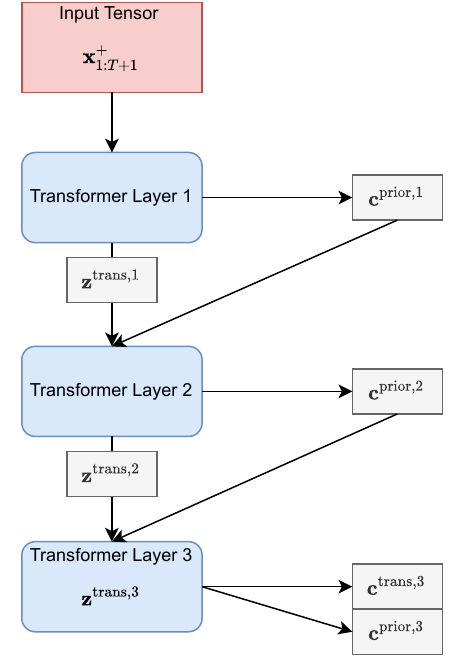}
    \caption{Diagram of the Deep Credibility Transformer architecture
    \eqref{Deep Credibility Transformer}.}
    \label{fig:deep_tf_diag}
\end{figure}

\end{document}